\documentclass[preprint,12pt]{elsarticle}

\usepackage{graphicx}
\usepackage{tabularx}
\usepackage{amssymb}
\usepackage{algorithm}
\usepackage{algpseudocode}
\usepackage{bold-extra}

\journal{Artificial Intelligence}

\begin{document}

\begin{frontmatter}
	
\title{XTE: Explainable Text Entailment}

\author[1]{Vivian S. Silva\corref{cor1}}
	\ead{vivian.santossilva@uni-passau.de}
\author[2]{Andr\'{e} Freitas}
	\ead{andre.freitas@manchester.ac.uk}
\author[1,3]{Siegfried Handschuh}
	\ead{siegfried.handschuh@unisg.ch}

\cortext[cor1]{Corresponding author}

\address[1]{Department of Computer Science and Mathematics, University of Passau, Germany}
\address[2]{School of Computer Science, University of Manchester, UK}
\address[3]{Institute of Computer Science,  University of St. Gallen, Switzerland}
	
\begin{abstract}
Text entailment, the task of determining whether a piece of text logically follows from another piece of text, is a key component in NLP, providing input for many semantic applications such as question answering, text summarization, information extraction, and machine translation, among others. Entailment scenarios can range from a simple syntactic variation to more complex semantic relationships between pieces of text, but most approaches try a one-size-fits-all solution that usually favors some scenario to the detriment of another. Furthermore, for entailments requiring world knowledge, most systems still work as a ``black box'', providing a yes/no answer that does not explain the underlying reasoning process. In this work, we introduce XTE -- Explainable Text Entailment -- a novel composite approach for recognizing text entailment which analyzes the entailment pair to decide whether it must be resolved syntactically or semantically. Also, if a semantic matching is involved, we make the answer interpretable, using external knowledge bases composed of structured lexical definitions to generate natural language justifications that explain the semantic relationship holding between the pieces of text. Besides outperforming well-established entailment algorithms, our composite approach gives an important step towards Explainable AI, allowing the inference model interpretation, making the semantic reasoning process explicit and understandable.
\end{abstract}


\end{frontmatter}
	
\section{Introduction}\label{sec: intro}
Many Natural Language Processing tasks, such as question answering, text summarization, information retrieval, and machine translation, among others, need to deal with language variability. Inputs and outputs can be expressed in different forms, and determining whether such forms are equivalent, that is, if a variant can be \textit{inferred} from the standard input or output, is an important task by itself. However, inference mechanisms built within specific NLP applications used to cover only the application's needs and could not be easily reused by other applications, since researchers in one area might not be aware of relevant methods developed in the context of another application \citep{dagan2013recognizing}. The text entailment paradigm was established as a unifying framework for applied inference, providing a means of delivering other NLP task from handling inference issues in an ad-hoc manner, using instead the outputs of an inference-dedicated mechanism.

Text entailment is defined as a directional relationship between a pair of text expressions, denoted by \textit{T} -- the entailing text, and \textit{H} -- the entailed hypothesis. We say that \textit{T} entails \textit{H} if, typically, a human reading \textit{T} would infer that \textit{H} is most likely true \citep{dagan2006pascal}. Three different scenarios can be observed: \\

\noindent (1) T and H are equivalent statements but expressed in different ways. Ex.: \\[5pt]
\noindent T: The badger is burrowing a hole. \\
H: A hole is being burrowed by the badger. \\

\noindent (2) H generalizes information from T. Ex.: \\[5pt]
\noindent T: A dog is riding a skateboard. \\
H: An animal is riding a skateboard. \\

\noindent (3) H present new information derived from T. Ex.: \\[5pt]
\noindent T: Iran is a signatory to the Chemical Weapons Convention. \\
H: The Chemical Weapons Convention is an agreement. \\

While (1) can usually be resolved syntactically, given that only the sentence structure is altered, and (2) requires only shallow semantic information, such as synonyms and hypernyms, (3) requires knowledge that goes beyond what is expressed in T and H, demanding the use of external world knowledge to solve the entailment.

Some text entailment approaches focus on exploring the syntactic structures of T and H, trying to transform the syntactic representation of T into that of H to determine whether they are equivalent and confirm the entailment. This kind of approach can fall short of identifying more complex semantic variations, like that observed in (3). On the other hand, techniques concentrating purely on finding semantic relations between T and H will struggle to deal with pairs like the one shown in (1) where only a syntactic variation holds. 

To overcome these issues and better address the variety of entailment phenomena, we propose the use of different methods to tackle different scenarios, integrated as components into a composite approach that performs a \textit{routing}, that is, it analyzes the entailment pair, identifies the most relevant phenomenon present, and sends it to the most suitable component to solve it.

We split the entailment phenomena into two broad categories: \textit{syntactic} and \textit{semantic}. For solving syntactic entailments, we adopt a tree edit distance algorithm, which operates over a dependency tree representation of T and H. For semantic entailments, we look for the semantic relationships holding between T and H, employing a distributional (word embedding-based) navigation algorithm that explores a graph knowledge base composed of natural language dictionary definitions. By finding paths is this graph linking T and H, we can provide human-readable justifications that shows explicitly what the semantic relationship holding between them is. Given the increasing importance of Explainable AI \citep{gunning2017explainable}, the ability to explain how decisions are reached is becoming a key demand for intelligent systems, and generating natural language justifications is an important feature for meeting this requirement and rendering a system \textit{interpretable}.

The contributions of this work, intended to equally address and balance the benefits for both the NLP and Explainable AI fields, are: 
\begin{itemize}
\item  a more flexible way to deal with different entailment scenarios, employing the most suitable method for each entailment phenomenon;
\item an interpretable definition-based commonsense reasoning model which, through the generation of natural language explanations, allows the final users to understand and assess the inference process leading to a decision;
\item a quantitative and qualitative analysis of different knowledge bases generated from various lexical resources, showing how they compare especially from the interpretability point of view.
\end{itemize}

\section{Related Work}\label{sec:relwork}
The introduction of the RTE (Recognizing Text Entailments) Challenges\footnote{https://goo.gl/R9zVqp} stimulated the development of a large number of text entailment frameworks. Starting in 2005, the RTE Challenges encouraged the creation of systems capable of capturing textual inferences, and, given the low accuracy achieved by the first participants, showed that much improvement was still required in the area \citep{ghuge2014survey}.

The first text entailment systems were mainly based on \textit{shallow methods}, relying only on word overlap and statistical lexical relations \citep{glickman2005probabilistic,perez2005application,newman2005ucd}. Later approaches have moved to more sophisticated approaches, or \textit{deep methods}, combining the analysis of the sentence structure with logical features and linguistic resources like WordNet, Framenet, and Verbnet, which can add some shallow semantic information to the syntactic data \citep{maccartney2008phrase,wang2008accuracy,sammons2009relation,harmeling2009inferring,stern2011confidence}. Edit distance \citep{kouylekov2005recognizing}, alignment \citep{wang2008divide} and transformation \citep{zanoli2016transformation} are some examples of such approaches. As a common starting point, they translate the text and hypothesis to some kind of (syntactic or semantic) representation and then try to determine if the representation of the hypothesis is subsumed by that of the text. 

Machine learning classification techniques are also employed \citep{jimenez2014unal,zhao2014ecnu,zhang2017context}, where T and H are represented as feature vectors, and multiple similarity measures (computed over lexical, syntactic and shallow semantic representations) are used to train a supervised machine learning model. Regarding the use of external world knowledge, \citet{silva2018recognizing} proposed an approach that focuses on semantic entailments, also exploring structured knowledge bases to find semantic relationships that confirm and explain the entailment.

Regarding the generation of natural language justifications, only a few text entailment systems \citep{clark2009inference,raina2005robust,silva2018recognizing} provide such feature; most approaches only output a yes/no answer and, sometimes, a confidence score, but no explanation or evidence that support the entailment decision. The Third RTE Challenge proposed an optional task which required a system to make three-way entailment decisions (\textit{entails}, \textit{contradicts}, \textit{neither}) and to justify its response \citep{voorhees2008contradictions}. Human evaluators judging the outputs provided by the competing systems pointed out a number of problems, notably the use of vague and abstract phrases such as ``there is a relation between'' and ``there is a match'', which shows that, rather than simply detecting that the semantic relation exists, it is also very important to describe what exactly this specific semantic relation is, so users can understand and trust the system's decisions.

With respect to the entailment recognition methods adopted in this work, although tree edit distance has already been used in text entailment \citep{kouylekov2005recognizing}, graph traversal approaches have been explored more often in other areas such as information retrieval \citep{frisse1988searching,gudivada1997information}, text mining \citep{aggarwal2011text}, text summarization \citep{ganesan2010opinosis}, and semantic similarity \citep{paul2016efficient}. Graphs were also already used in text entailment \citep{kotlerman2015textual}, but as a set of potential entailment rules extracted from text. The use of graph traversal methods for exploring independent, external knowledge bases for injecting world knowledge in the entailment recognition process and for generating explanations for a system's decisions is still an emerging field.

\section{Text Entailment vs. Natural Language Inference}
In the last years, with the end of the RTE Challenges, the development of new textual entailment approaches have slowed down. On the other hand, a new subtask derived from it have emerged, leveraged by a new set of datasets and machine learning methods. As opposed to the original text entailment task, which is a binary classification task where the answer is either \textit{yes} or \textit{no} (corresponding to \textit{entailment} or \textit{non-entailment}), the \textit{Natural Language Inference} (NLI) subtask is intended to perform a three-way classification, labeling entailment pairs as \textit{entailment}, \textit{neutral}, or \textit{contradiction}.

NLI is usually associated with deep learning methods, which was enabled by the introduction of large machine learning oriented datasets, such as the \textit{Stanford Natural Language Inference} (SNLI) corpus \citep{bowman2015large}, and the \textit{Multi-Genre Natural Language Inference} (MultiNLI) corpus \citep{williams2018broad}. Benefiting from the large amount of training data, NLI systems can make use of models and techniques now widely adopted for natural language text processing, among which stand out the \textit{Long Short Term Memory} (LSTM) models and the attention mechanisms integrated into deep neural networks.

In the NLI task, the text (T) sentence is usually called the \textit{premise}, and sentence embedding is a common way of representing the premise and the hypothesis. LSTM models, be it plain \citep{bowman2015large}, or enriched with attention-weighted representations \citep{rocktaschel2016reasoning}, word-by-word matching between sentences \citep{wang2016learning}, Bidirectional LSTM \citep{chen2017enhanced}, or a decomposable attention model \citep{parikh2016decomposable}, are the most common architectures. A number of variants regarding the attention mechanism have also been proposed, including \textit{inner-attention} to detect the most important portions of one sentence in the pair regardless of the content of the other one \citep{liu2016learning}, \textit{self-attention} to model the long-term dependencies in a sentence \citep{im2017distance,gong2017natural}, and \textit{co-attention} to preserve information from all the network layers \citep{kim2019semantic}. Despite the differences in the way they attend sentences and align their words, these approaches build upon similar architectures, using LSTM or BiLSTM models with no or almost no feature engineering, and no external resources.

Even though NLI datasets are semantically simpler than text entailment ones, still not all the knowledge needed for the inference is self-contained within the training data. Some approaches try to address this issue by incorporating external knowledge in the inference process. \citet{chen2018neural} do this by extracting synonymy, antonymy, hypernymy, and co-hyponymy (relation between sibling words, that is, words having the same hypernym) relations from WordNet. \citet{wang2019improving} goes further and, besides WordNet, use also ConceptNet and DBPedia as knowledge sources.

Although NLI systems show great quantitative improvement when compared with text entailment applications, since all of them use very advanced models, the increasing number of different approaches present only incremental improvements among them. Furthermore, even though the advances introduced in the NLI field are arguably invaluable, the high accuracy the approaches achieve may nevertheless be partly influenced by bias in the training datasets. In a study conducted by \citet{gururangan2018annotation} (in which Samuel R. Bowman, one of the researchers responsible for the creation of both SNLI and MultiNLI, also participated), it is shown that NLI datasets contain a significant number of annotation artifacts that can help a classifier detect the correct class without ever observing the premise. The presence of such artifacts is a result of the crowdsourcing process adopted for the dataset creation, because crowd workers adopt heuristics in order to generate hypotheses quickly and efficiently, producing certain patterns in the data. Through a shallow statistical analysis of the data, focusing on lexical choice and sentence length, they found, for example, that entailed hypotheses tend to contain gender-neutral references to people, purpose clauses are a sign of neutral hypotheses, and negation is correlated with contradiction.

Besides the dataset statistical analysis, they also built a hypothesis-only classifier, showing that a significant portion of SNLI and MultiNLI test sets can be correctly classified without looking at the premise. Then, they re-evaluated high-performing NLI models on the subset of examples on which the hypothesis-only classifier failed (which were considered to be ``hard''), showing that the performance of these models on the ``hard'' subset is dramatically lower than their performance on the rest of the instances. They conclude that supervised models perform well on these datasets without actually modeling natural language inference because they leverage annotation artifacts and these artifacts inflate model performance, so the success of NLI models to date has been overestimated. \citet{poliak2018hypothesis} reinforces these conclusions, implementing a similar hypothesis-only classifier but extending the study to other eight datasets besides SNLI and MultiNLI, underlining that such statistical irregularities lead models to skimp over a fundamental principle of textual entailment and, by extension, of NLI: that the truth of the hypothesis necessarily follows from the premise and, then, the premise must be indispensable if actual inference is to be performed.

The problems evidenced by the bias in the datasets show that NLI is still an open challenge, but one more issue can be highlighted: due to their increasingly more complex architectures, NLI models will invariably show poor interpretability, making it even harder for users to know how decisions are reached, and, therefore, if they are reliable or not. As deep neural network models, they are not transparent, and commonly do not provide post-hoc explanations. Regarding explanations, the recently released e-SNLI dataset \citep{camburu2018snli}, an extension of SNLI with human-annotated natural language explanations for each premise-hypothesis pair, can leverage the developments in this area. So far, a single approach \citep{thorne2019generating} has used this dataset, and only for token-level explanation, that is, only the tokens in the premise and in the hypothesis that are relevant for the inference are presented (which is basically the output of the attention mechanism). Fully human-readable explanations made up of full concise and connected natural language sentences are yet to be addressed in NLI.

\section{Syntactic-Semantic Composite Text Entailment}\label{sec:approach}
This work builds upon a previous version of the approach presented in \citep{silva2019exploring}, extending the entailment framework to consider context information in order to better capture the semantics of the sentences as a whole, and also performing more extensive experiments, therefore, introducing novel and enhanced results. The central point is the notion that text entailment can involve syntactic or semantic phenomena, and each of these phenomena categories requires specific approaches to be solved. In the first case, an analysis of the syntactic structure of the sentences may be enough, while in the second it is necessary to identify the semantic relationship holding between the text and the hypothesis. On the other hand, looking for semantic relationships where only a syntactic variation occurs or comparing syntactic structures of very (syntactically) different sentences can be highly counterproductive, hence the importance of choosing the suitable method first and foremost.

To pick the best approach, we need to answer the following question: \textit{Can there be a semantic relationship between T and H?} We assume that a semantic relationship must hold between two entities $e_{1}$ and $e_{2}$, $e_{1} \neq e_{2}$, both referring to a third entity, which we call the \textit{referent}, or $r$. 

The routing mechanism that will check these conditions relies on the notion of \textit{overlap} between the text and the hypothesis. The overlap $O$ is computed over the bag-of-words representation of T and H, denoted by:

\begin{equation}\label{eq:tset}
T' = \{t_{1}, t_{2}, ..., t_{n}\}
\end{equation}

\noindent where $t_{i}$ are tokens in T and $n$ is the size of $T'$, and

\begin{equation}\label{eq:hset}
H' = \{h_{1}, h_{2}, ..., h_{m}\}
\end{equation}

\noindent where $h_{i}$ are tokens in H and $m$ is the size of $H'$. Therefore:

\begin{equation}\label{eq:oset}
O = T' \cap H' = \{w_{1}, w_{2}, ..., w_{k}\}
\end{equation}

\noindent where $k$ is the size of $O$. Formalizing the aforementioned conditions for the existence of a semantic relationship between T and H, we have that:

\begin{equation}\label{eq:cond1}
\exists e_{1} \in T' \wedge \exists e_{2} \in H' \wedge  e_{1} \neq e_{2}
\end{equation}

\begin{equation}\label{eq:cond2}
\exists r \in O
\end{equation}

After computing $O$, three scenarios may occur: \\

\noindent(1) \textit{total overlap}, where all the tokens of $H'$ are contained in $T'$ or (less commonly) vice-versa, that is, $k = m$ or $k = n$. In this case, the condition \ref{eq:cond1} is not satisfied; \\[5pt]
(2) \textit{partial overlap}, where some but not all of the of tokens of $T'$ are contained in $H'$, so $k < n$ and $k < m$. Both conditions \ref{eq:cond1} and \ref{eq:cond2} are met in this scenario; \\[5pt]
(3) \textit{null overlap}, that is, no tokens of $T'$ are contained in $H'$, so $O = \varnothing$ and $k = 0$. Since $O$ is empty, the condition \ref{eq:cond2} can't be satisfied. \\[5pt]

Given that we can look for a semantic relationship between T and H solely when both conditions \ref{eq:cond1} and \ref{eq:cond2} are met, the entailment pair will be solved semantically only when a partial overlap occurs. Otherwise, the pair will be solved syntactically because, if there is a total overlap, there are no entities $e_{1}$ and $e_{2}$ such that $e_{1} \neq e_{2}$ for which a semantic relationship may hold, and, in the case of a null overlap, there is no referent $r$, so, even if there are some potential candidates $e_{1}$ and $e_{2}$ that could be semantically related, it is more likely (although not certain) that they are referring to completely different entities.

\begin{figure*}[ht]
	\centering
	\includegraphics[width=\textwidth]{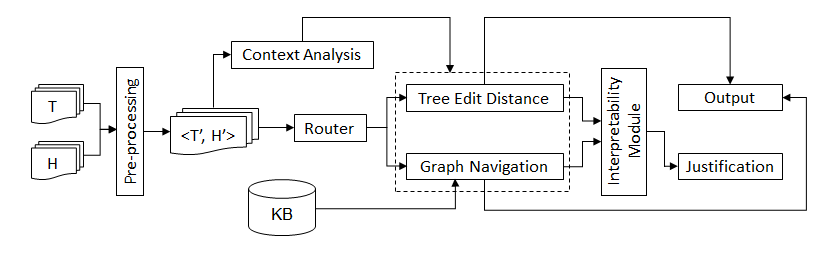}
	\caption{General architecture of XTE (Explainable Text Entailment)}
	\label{fig:architechture}
\end{figure*}

For solving entailments syntactically, we use the \textbf{Tree Edit Distance} (TED) model, and for dealing with entailments involving semantic phenomena we employ the \textbf{Distributional Graph Navigation} (DGN) model. An additional \textbf{Context Analysis} module feeds both models with extra information extracted from the entailment pair. Following a preprocessing stage that generates $T'$ and $H'$, the router computes $O$ and sends the entailment pair either to the TED or to the DGN model, according to the aforementioned conditions. After the entailment is solved by the suitable model, returning \textit{yes} or \textit{no} as the output, an interpretability module uses the evidence produced by the entailment algorithm to generate a natural language justification explaining the algorithm's decision. The general architecture of the XTE (Explainable Text Entailment) approach is shown in Figure \ref{fig:architechture}.

\subsection{Tree Edit Distance}\label{sec:ted}
\begin{figure*}[ht]
	\centering
	\includegraphics[width=\textwidth]{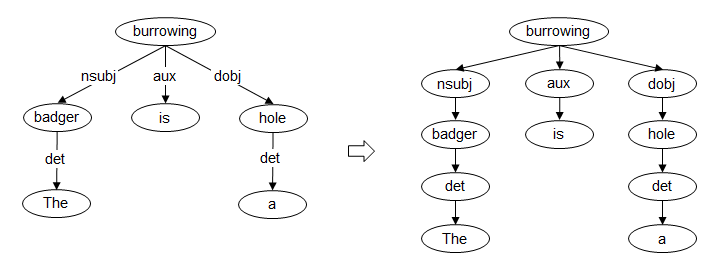}
	\caption{Dependency graph (left) and the resulting dependency tree (right) which is sent to the tree edit distance algorithm}
	\label{fig:dep_tree}
\end{figure*}

The Tree Edit Distance algorithm computes the minimal-cost sequence of operations, namely \textit{insertion}, \textit{deletion} and \textit{replacement} of nodes, necessary to transform the tree representation of T into the tree that represents H. We use the \textit{All Paths Tree Edit Distance} (APTED) \citep{pawlik2016tree}, which improves over the classical algorithm of \citet{zhang1989simple} by being tree-shape independent. The edit distance is computed over the syntactic dependency trees of T and H, generated by the Stanford dependency parser \citep{chen2014fast}. This parser generates a dependency graph, but it can be easily converted to an acyclic tree, where nodes with more than one incoming edge are expanded only at the first time they are referenced, and represented as childless nodes in subsequent references (similar to the pretty-print string representation provided by the parser for the original graph). We represent dependencies between terms, which are labeled edges in the original graph, as intermediary nodes between the two nodes they link, that is, the two dependency's arguments. Figure \ref{fig:dep_tree} shows the graph generated by the dependency parser for the text sentence T in the example (1) in Section \ref{sec: intro},  ``The badger is burrowing a hole'', and the resulting dependency tree which will be sent as input to the TED algorithm.

Given that we represent dependencies as nodes in the tree, our TED model penalizes node replacement more than insertion and deletion, because replacing a node $x$ between nodes $a$ and $b$ in T by a node $y$ between the same nodes $a$ and $b$ in H means changing the dependency between them, or changing one of the arguments of a dependency, if the replacement comes before or after a sequence of two nodes $a$ and $b$ which are identical in T and H. This is done by a \textit{weighted cost model} with higher weight for replacements than for insertions and deletions, and by the calculation of the relative edit distance \textit{relDist}, which is the edit distance $dist$ relative to the difference $diff$ between the sizes of the two trees, given by $relDist = dist/diff$. If the two trees are roughly the same size, but many edit operations are performed, they are probably replacements, which means many dependencies and/or arguments are being changed, so $diff$ is low and $relDist$ increases. On the other hand, if approximately the same number of operations are performed for trees having different sizes (usually, T larger than H), there will be more insertions and/or deletions. In this case, $diff$ is higher and $relDist$ decreases, which favors scenarios where the tree for H is a subtree of the tree for T, and, therefore, insertions/deletions will occur more often and affect the validity of the entailment less than replacements. The $relDist$ is then compared against a threshold $t$, and the pair is classified as an entailment if $relDist < t$, and as a non-entailment otherwise.

\subsection{Distributional Graph Navigation}\label{sec:dgn}
The Distributional Graph Navigation model is based on two main pillars: the use of a knowledge graph automatically extracted from natural language lexical definitions as a world knowledge base, and a navigation mechanism based on distributional semantics to traverse this graph and find paths between the text and the hypothesis. A path between T and H explains the semantic relationships holding between them, confirming the entailment (by showing that a relationship indeed exists) while also providing evidence that it is true (by enabling the generation of an explanation from its nodes contents).

\subsubsection{The Definition Knowledge Graph}\label{sec:dkg}
Generating natural language justifications is an important feature for increasing a system's interpretability, and the generation of such explanations can be leveraged by the use of external sources of world knowledge. Dictionary-style definitions are a rich source of such knowledge and, different from formal, structured resources like ontologies, they are domain-independent and largely available. Many NLP systems, including text entailment systems \citep{clark2008using,herrera2006textual}, already explore lexicons, among which the most used is WordNet \citep{fellbaum1998wordnet}, but they usually look only at its structured information, that is, links such as synonyms, hypernyms, etc. The natural language definitions are left aside, although they contain the largest amount of relevant information about an entity: its type, essential attributes, primary functions, and often many non-essential, but very informative, attributes as well.

We rely on the knowledge provided by lexical dictionary definitions for looking for relationships between the text and the hypothesis whenever the entailment is solved semantically. To make use of natural language definitions in our approach, we structure them, converting a whole dictionary into a knowledge graph following the conceptual representation model proposed by \citet{silva2016categorization}. In this model, the definitions are split into entity-centered \textit{semantic roles}. Differently from the commonly used event-centered semantic roles, which define the semantic relations holding among a predicate (the main verb in a clause) and its associated participants and properties \citep{marquez2008semantic}, definition's semantic roles express the part played by an expression in a definition, showing how it relates to the \textit{definiendum}, that is, the entity being defined. Table \ref{tab:roles} lists the semantic roles for lexical definitions present in the model.

\begin{table}[ht]
	\centering
	\begin{tabularx}{\linewidth}{l X}
		\textbf{Role}        & \textbf{Description}                                                                                                                                  \\ \hline
		Supertype            & the immediate or ancestral entity's superclass                                                                                                        \\
		Differentia quality  & a quality that distinguishes the entity from the others under the same supertype                                                                      \\
		Differentia event    & an event (action, state or process) in which the entity participates and that is mandatory to distinguish it from the others under the same supertype \\
		Event location       & the location of a differentia event                                                                                                                   \\
		Event time           & the time in which a differentia event happens                                                                                                         \\
		Origin location      & the entity's location of origin                                                                                                                       \\
		Quality modifier     & degree, frequency or manner modifiers that constrain a differentia quality                                                                            \\
		Purpose              & the main goal of the entity's existence or occurrence                                                                                                 \\
		Associated fact      & a fact whose occurrence is/was linked to the entity's existence or occurrence                                                                         \\
		Accessory determiner & a determiner expression that doesn't constrain the supertype-differentia scope                                                                        \\
		Accessory quality    & a quality that is not essential to characterize the entity                                                                                            \\
		{[}\textit{Role}{]} particle  & a particle, such as a phrasal verb complement, non-contiguous to the other role components                                                           
	\end{tabularx}
	\caption{Semantic roles for dictionary definitions}
	\label{tab:roles}
\end{table}

This model allows a structured semantic representation of natural language definitions and enables the selection of the portions of information that are relevant for a given reasoning task. For building the definition knowledge graph (DKG), we followed the methodology introduced in \citep{silva2018building}, which comprises the following steps:

\noindent\textbf{Definitions sample selection:} we selected a random sample of noun and verb definitions to be annotated so we could train a supervised machine learning model to classify the data. We used the glosses extracted from the WordNet database, from which we randomly selected 4,000 definitions, being 3,443 noun definitions and 557 verb definitions (the verb database size in WordNet is around 17\% of the noun database size).

\noindent\textbf{Automatic pre-annotation:} using the syntactic patterns identified by statistical analysis described by \citet{silva2016categorization}, we implemented a rule-based heuristic to automatically pre-annotate the set of 4,000 definitions. This heuristic links each phrasal node in a sentence's syntactic tree to the semantic role most often associated with it. For example, the \textit{supertype} for a noun definition is usually the innermost and leftmost noun phrase (NP) that contains at least one noun (NN); a \textit{differentia event} is usually either a subordinate clause (SBAR) or a verb phrase (VP); and so on. The syntactic parse trees for each definition were generated and queried with the aid of the Stanford parser \citep{manning2014stanford}.

\noindent\textbf{Data curation:} after the automatic pre-annotation, the definitions were manually curated so misclassifications were fixed and segments missing a role were assigned the appropriate one. The manual data curation ensured that the whole definition was consistently classified, that is, that every segment was associated with the most suitable semantic role label.

\noindent\textbf{Classifier training:} the curated dataset was then used to train a Recurrent Neural Network (RNN) machine learning model designed for sequence labeling. We used the RNN implementation provided by \citet{mesnil2015using}, which reports state-of-the-art results for the slot filling task. The dataset was split into training (68\%), validation (17\%) and test (15\%) sets. The best accuracy reached during training was of 77.24\%.

\noindent\textbf{Database classification:} the trained classifier was then used to label the whole set of definitions from different lexical resources, as detailed in Section \ref{sec:kb}. Each resource gave origin to a different DKG, which allowed us to test our approach with different configurations and compare the resources among them (see Section \ref{sec:comparison}).

\noindent\textbf{Data post-processing:} we then passed the classified data through a post-processing phase aimed at fixing classification that missed the supertype role. This role is a mandatory component in a well-formed definition and, as detailed in the next step, the RDF model is structured around it. Missing supertypes were identified according to the same syntactic rules used for pre-annotation, keeping the remaining classification unchanged.

\begin{figure}[ht]
	\centering
	\includegraphics[width=\textwidth]{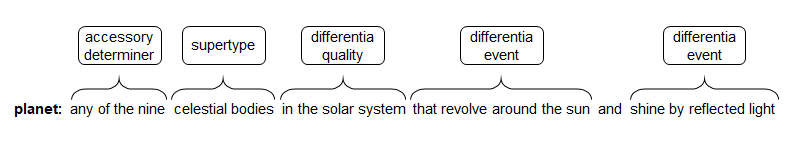}
	\caption{The labeled definition for the concept ``planet''}
	\label{fig:labeled_def}
\end{figure}

\begin{figure}[ht]
	\centering
	\includegraphics[width=3.3in]{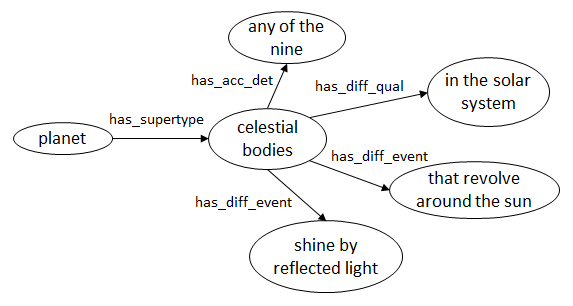}
	\caption{The graph representation of a lexical definition. The node labeled ``planet'' is an entity node (the entity being defined), and all the other ones are role nodes}
	\label{fig:graph}
\end{figure}

\noindent\textbf{RDF conversion:} finally, the labeled definitions were serialized in RDF format. In the RDF graph, the entity being defined is a node (the \textit{entity node}), and each semantic role in its definition is another role (the \textit{role nodes}). The entity node is linked to its supertype role node, which is, in turn, linked to all the other role nodes. Consider, for example, the definition (from WordNet) for the concept ``planet'': ``any of the nine large celestial bodies in the solar system that revolve around the sun and shine by reflected light''. The output of the labeling stage is depicted in Figure \ref{fig:labeled_def}, and the resulting RDF subgraph is shown in Figure \ref{fig:graph}.

When using this graph to recognize semantic text entailments, we assume that whenever we can find a path between two entities $e_{1}$ and $e_{2}$, $e_{1} \in T'$ and $e_{2} \in H'$ (see condition \ref{eq:cond1} in Section \ref{sec:approach}), the entailment is true, and this path is used to justify the entailment, showing explicitly what the relationship between the entities is.

\subsubsection{The Distributional Navigation Algorithm}\label{sec:dna}
Distributional Semantic Models (DSMs) are grounded in the distributional hypothesis, which states that words that occur in similar contexts tend to have similar meanings \citep{turney2010frequency}. DSMs allow the approximation of a word meaning representing it as a vector summarizing its pattern of co-occurrence in large text corpora \citep{marelli2014sick}.

DSMs can be used to compute the \textit{semantic similarity/relatedness} measure between words. This computation is used as a heuristic to navigate in a graph knowledge base in the approach proposed by \citet{freitas2014distributional}, where they define the \textit{Distributional Navigation Algorithm} (DNA), which corresponds to a selective reasoning process in the knowledge graph. Given a pair of terms, namely a \textit{source} and a \textit{target}, and a threshold $\eta$, the DNA finds all paths from source to target, with length $l$, formed by concepts semantically related to target wrt $\eta$ \citep{freitas2014distributional}.

In the text entailment context, and, more specifically, for semantic entailments, the source and target are the entities $e_{1} \in T'$ and $e_{2} \in H'$, $e_{1} \neq e_{2}$, which we assume have some kind of semantic relationship between them. A path in the definition knowledge graph linking these entities, then, explains what this relationship is, confirming the entailment, or rejecting it in case no path is found.

\begin{figure}[ht]
	\centering
	\includegraphics[]{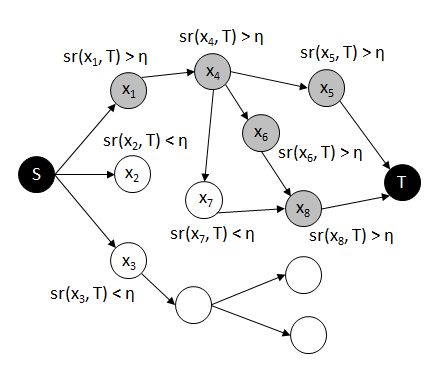}
	\caption{The distributional navigation algorithm. Gray nodes, for which $sr(x_{i}, T) > \eta$, make up valid paths between the source node S and the target node T. The path $\{S, x_{1}, x_{4},  x_{5}, T\}$ is the shortest one}
	\label{fig:dna}
\end{figure}

We implement the DNA as a \textit{depth-first search} algorithm, exploring first the paths whose next node to be visited has the highest semantic similarity value wrt the target. Given a node in the DKG, starting from the source $S$, the algorithm retrieves all its neighbors $\{x_{1}, x_{2}, ..., x_{n}\}$ and computes the similarity relatedness $sr(x_{i}, target)$, keeping only the nodes for which $sr > \eta$ in the set of nodes to be visited next. Each of these nodes generates a new path, and, for each path, the search goes on until the next node to be visited is equal to the target or a synonym of it, or until the maximum path length is reached. If no path reaches the target before the maximum number of paths is reached, the search stops. The distributional graph navigation mechanism is schematized in Figure \ref{fig:dna}. The DGN algorithm, which takes as inputs a definition knowledge graph $G$, a source word $S$, a target word $T$, a threshold $\eta$, a maximum path length $l$, and a maximum number of paths $m$, and outputs the set $P$ of paths from $S$ to $T$, is listed in Algorithm \ref{alg:dgn}.

\begin{algorithm}[]
	{\footnotesize
	\caption{Distributional Graph Navigation Algorithm}
	\label{alg:dgn}
	\begin{algorithmic}[1]
		\Procedure{DGN}{$G, S, T, \eta, l, m$}
			\State $P \gets \varnothing$
			\State $stack \gets \varnothing$
			\State $newPath \gets [S]$	
			\State \textbf{Push}($stack, newPath$) \Comment{adds the $newPath$ to the $stack$}
			\While {$stack \neq \varnothing$ \textbf{and} $P.size < m$}
				\State $path \gets \textbf{Pop}(stack)$ \Comment{pulls the path at the top of the $stack$}
				\State $nextNode \gets path.lastNode$
				\While {$nextNode \neq T$ \textbf{and} $path.length < l$}
					\State $entityNodes \gets \varnothing$
					\ForAll {$e_{i} \in G$}
						\If {$e_{i} = nextNode$} \label{line:ent_search}
							\State \textbf{Add}($entityNodes, e_{i}$)
						\EndIf	
					\EndFor
					\State $roleNodes \gets \varnothing$
					\State $bestRoles \gets \varnothing$
					\ForAll {$e_{i} \in entityNodes$}
						\State \textbf{Add}($roleNodes$, \textbf{Neighbors}($e_{i}$)) \label{line:neighbors}
					\EndFor
					\ForAll {$r_{i} \in roleNodes$}
						\If {$sr(r_{i}, T) > \eta$} \label{line:filter}
							\State \textbf{Add}($bestRoles, r_{i}$)
						\EndIf	
					\EndFor	
					\State $bestRoles \gets \textbf{Sort}(bestRoles)$
					\State $nextNodes \gets \varnothing$
					\ForAll {$b_{i} \in bestRoles$}
						\State \textbf{Add}($nextNodes$, \textbf{HeadWords}($b_{i}$))
					\EndFor
					\State $nextNodes \gets \textbf{Sort}(nextNodes)$
					\For {$x_{i} \in nextNodes, i \leftarrow 2, n$}
						\State $newPath \gets path$
						\State \textbf{Add}($newPath, x_{i}$)
						\State \textbf{Push}($stack, newPath$)
					\EndFor
					\State $nextNode \gets x_{1}$
					\State \textbf{Add}($path, nextNode$)
					\If {$nextNode = T$ \textbf{or} \textbf{Synonyms}($nextNode, T$)}\label{line:syn}
						\State \textbf{Add}($P, path$)
					\EndIf
				\EndWhile
			\EndWhile
			\State \textbf{return} $P$
		\EndProcedure
	\end{algorithmic}}
\end{algorithm}

Depending on the lexical resource from which the definitions are extracted, entity nodes in a DKG can be identified by a single word or phrase, or by a \textit{synset}, that is, a set of synonym words or phrases. Starting from the source word $S$, the DGN retrieves all entity nodes identified by $S$ or having $S$ as one of the words in its identifying synset (line \ref{line:ent_search}). Then it retrieves all the neighbors of each entity node, that is, the role nodes that make up its definition (line \ref{line:neighbors}), and keeps only the best ones (line \ref{line:filter}). 

The next nodes to be visited are given by words present in a role node, which we call the \textit{head words}. For getting the head words, we first remove all stop words and words with low \textit{inverse document frequency} (IDF), which are words that occur too frequently (for example, verbs such as ``get'', ``put'', ``cause'' or ``make'') and can be reached from almost any node in the graph, leading to diverting paths. IDF is calculated using as the corpus the same linguistic resource that gave origin to the knowledge graph being explored by the algorithm. After removing the irrelevant words, we compute the semantic similarity $sr$ between each remaining word and the target word $T$, sort the results and keep only the top $k$ words. The highest scoring head word will be the next node to be visited, that is, the next entity node to be searched, and all the other head words are added to a copy of the current path, generating a new path which will be pushed to the stack to be explored later.

As mentioned earlier, the search stops successfully when the next node to be visited is equal to the target, but also when it is one of the target's synonyms (line \ref{line:syn}). This is done with the aid of a synonym table, built from synonym lists gathered across all the tested lexical resources (see Section \ref{sec:kb}) and other online resources\footnote{https://bit.ly/2VPkywz, https://bit.ly/2kimBWS}.

Word sense disambiguation comes as a natural consequence of the distributional navigation mechanism while choosing the next nodes to be visited in the graph: by looking for the word/phrases that are more semantically related to the target $T$, the algorithm naturally selects the correct (or at least the closest) word senses, since unrelated word meanings will have lower similarity scores wrt the target, and the paths containing them will be excluded by the algorithm.

According to \citet{freitas2014distributional}, the worst-case time complexity of the algorithm implemented as a depth-first search ``is $O(b^{l})$, where $b$ is the branching factor and $l$ is the depth limit''. They show that the algorithm's selectivity ensures that the number of paths does not grow exponentially even when the depth limit increases. In our implementation, the maximum number of paths and the maximum path length (depth limit) were obtained empirically in order to optimize the search. 

\subsubsection{Recognizing and Justifying Entailments through the DGN}\label{sec:recjust}
For applying the distributional graph navigation algorithm (Section \ref{sec:dgn}) over definition knowledge graphs (Section \ref{sec:dkg}) for recognizing and justifying semantic entailments, we first need to identify the source-target pairs, that is, the pairs of entities $\{e_{i}, e_{j}\}$ which will be sent as input to the DGN. Using the information from the sets $T'$ (Equation \ref{eq:tset}), $H'$ (Equation \ref{eq:hset}) and $O$ (Equation \ref{eq:oset}), we compute the sets $T''$ and $H''$, where:

\begin{equation}\label{eq:t2set}
T'' = T' - O
\end{equation}

\begin{equation}\label{eq:h2set}
H'' = H' - O
\end{equation}

Also using DSMs, we then compute the semantic similarity measures between $T''$ and $H''$ as the Cartesian product P:

\begin{equation}\label{eq:pset}
P = T'' \times H''
\end{equation}

The results are then sorted and the $k$ highest scoring pairs are selected, making up the set $P'$. Each pair $\{e_{i}, e_{j}\} \in P'$ is sent to the DGN algorithm, which returns a set of paths between $e_{i}$ and $e_{j}$. From the set of all paths returned for all pairs, we choose the shortest one, which is the one that offers the shortest distance between a source and a target and, therefore, shows that their meanings are more closely related. If no path is found at all, then the entailment is rejected.

A path in the DKG is composed by a sequence of entity nodes and the relevant role nodes linked to them (see Section \ref{sec:dkg}). This path is then sent to the interpretability module, which formats them into a human-readable justification, which shows what the relationship between the source (and therefore, the text T), and the target (the hypothesis H) is, making clear what was the reasoning followed by the algorithm. The procedure for recognizing an entailment through the DGN is listed in Algorithm \ref{alg:sementail} (for readability, further parameters for the DGN procedure were omitted in line \ref{line:calldgn}).

\begin{algorithm}[t]
	\caption{Semantic Entailment Recognition through the DGN Algorithm}\label{euclid}
	\label{alg:sementail}
	\begin{algorithmic}[1]
		\Procedure{ProcessEntailment}{$T, H$}
			\State $T' \gets$ \textbf{Tokenize}($T$)
			\State $H' \gets$ \textbf{Tokenize}($H$)
			\State $O \gets T' \cap H'$
			\State $T'' \gets T' - O$
			\State $H'' \gets H' - O$
			\State $P \gets T'' \times H''$
			\State $P' \gets \textbf{TopK}(\textbf{Sort}(P))$
			\ForAll {$\{e_{i}, e_{j}\} \in P'$}
				\State \textbf{Add}($allPaths, DGN(e_{i}, e_{j})$)\label{line:calldgn}
			\EndFor
			\If {$allPaths = \varnothing$}
				\State $entailment \gets false$
			\Else
				\State $entailment \gets true$
				\State $bestPath \gets \textbf{ShortestPath}(allPaths)$
				\State $justification \gets \textbf{WriteJustification}(bestPath)$
			\EndIf
			\State \textbf{return} $entailment, justification$
		\EndProcedure
	\end{algorithmic}
\end{algorithm}

To illustrate the steps followed by the DGN while traversing a knowledge graph to solve an entailment, consider the entailment pair 47.7 from the BPI\footnote{http://www.cs.utexas.edu/users/pclark/bpi-test-suite/} dataset: \\

\noindent 47.4 T: Iran is a signatory to the Chemical Weapons Convention. \\
47.4 H: The Chemical Weapons Convention is an agreement. \\

\begin{figure}[ht]
	\centering
	\includegraphics[width=\textwidth]{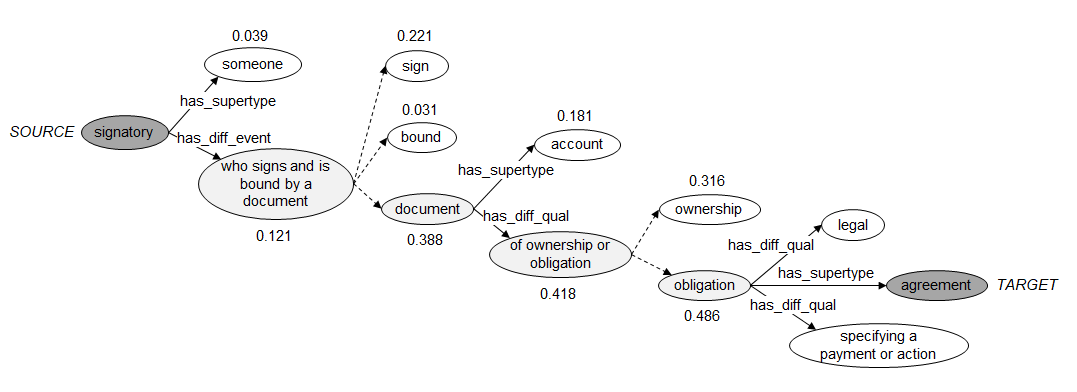}
	\caption{A path, indicated by the gray nodes, between source node ``signatory'' and target node ``agreement'' in a DKG. Full lines represent actual edges in the graph, while dashed lines represent the algorithm's internal operations, in this case the extraction of head words for multi-word expression nodes. Numbers show the semantic relatedness between each node and the target}
	\label{fig:path}
\end{figure}

In this example, the best source-target pair is $e_{1}$ = ``signatory'' and $e_{2}$ = ``agreement'', and the referent $r$ = ``Chemical Weapons Convention'', since both $e_{1}$ and $e_{2}$ refer to this concept. The best path between the source and the target in a DKG, as well as all the semantic similarity measures between each node retrieved by the algorithm and the target, are shown in Figure \ref{fig:path}.

In this path, nodes are linked either by the \textit{has\_supertype} property, which defines the \textit{kind} of an entity, or by the \textit{has\_diff\_event} or \textit{has\_diff\_qual} properties, which introduce \textit{qualifiers} for the entities they describe. In the second case the \textit{supertype} node is also included in the path, because \textit{differentia} role nodes (as well as almost all the other role nodes) don't make much sense without the supertype they refer to. Since the justification takes into account the content of the nodes and the relationships between them, that is, the role names (see Section \ref{sec:dkg}), the final, human-readable explanation generated by the algorithm from this sequence of nodes is: \\

\noindent \texttt{A \textbf{signatory} is someone who signs and is bound by a \textbf{document} \\
A \textbf{document} is an account of ownership or \textbf{obligation} \\
An \textbf{obligation} is a kind of \textbf{agreement}} \\

The natural language justifications are not a formal logical proof of the entailment but rather explanations based on commonsense, intended to be close to what a human being, employing their accumulated world knowledge, would use to justify the link between concepts.

\subsection{Context Analysis}\label{sec:context}
The goal of the Context Analysis module is to provide both the Tree Edit Distance and the Distributional Graph Navigation models with information they can't easily grasp, and that, if missed, can lead to erroneous conclusions. In the Tree Edit Distance case, this can happen when minimal, slight modifications completely changes the meaning of a sentence. Such modifications may yield only a small edit distance between T and H, resulting in a wrong \textit{entailment} classification for the pair.

For semantic entailments, the extraction of further contextual information is even more critical, since the Distributional Graph Navigation model, though looking for common referents, primarily considers pairs of terms in isolation, and not the sentences as a whole. This means that, even when there are two entities $e_{1} \in T'$ and $e_{2} \in H'$ with a high semantic relatedness score, there may also be, at another point in the sentences, contradictory or inconsistent information which invalidates the entailment, but that the DGN won't catch.

The Context Analysis module receives as input the tokenized output from the preprocessing stage, discards the overlapping entities in the set $O$, and, using syntactic and semantic features, analyzes the remaining words/phrases in order to look for the following phenomena\footnote{Examples from the SICK dataset (http://clic.cimec.unitn.it/composes/sick.html)}: \\

\noindent \textbf{Simple Negation:} T is a simple negation of H. Example: \\

\noindent 1127 T: A sea turtle is not hunting for fish \\
1127 H: A sea turtle is hunting for fish \\
1127 A: NO \\

Negation adverbs will mostly be considered as stop words, so entailment pairs where these are the only divergent words will be sent to the Tree Edit Distance model. Detecting the negation allows the model to classify the pair as a non-entailment even if the final edit distance is well below the threshold. \\

\noindent \textbf{Opposition:} T contains a term which is an antonym of a term in H. Example: \\

\noindent 3706 T: A woman is taking off eyeshadow \\
3706 H: A woman is putting make-up on \\
3706 A: NO \\

Detecting opposition is an important step when entailments are solved through the Distributional Navigation model. In the above example, the DGN would detect ``eyeshadow'' in T and ``make-up'' in H as a candidate source-target pair and most likely find a path in a DKG linking both entities (since ``eyeshadow'' is a kind of ``make-up''), but the presence of antonym terms ``take off'' in T and ``put'' in H prevents the pair from being misclassified as an entailment. Opposition detection is performed with the aid of an antonym table, built from antonym lists extracted from all the tested lexical resources (see Section \ref{sec:kb}) and other online resources\footnote{\raggedright https://bit.ly/2Uy3eMf, https://bit.ly/2J6N1wd, https://bit.ly/2HcDK3W, https://bit.ly/2O0GCBs, https://bit.ly/2Tv7UWN, https://bit.ly/2XSm5DN}. \\

\noindent \textbf{Inverse Specialization:} H specializes some information in T. Since text entailment is a directional relationship from T to H, specializations are valid only in this direction, not the other way around. Example: \\

\noindent 1382 T: A person is rinsing a steak with water \\
1382 H: A man is rinsing a large piece of meat \\
1382 A: NO \\

In this example, H specializes T since ``person'' is more general than ``man''. As much as opposition detection, inverse specialization detection plays an important part in preventing the DGN from misclassifying the entailment pair (in the above example, by find a relationship between ``steak'' and ``meat''). In the correct direction, that is, from T to H, specializations can be easily detected also by the DGN model, since they are a kind of semantic relationship; for detecting only inverse specializations we use the hypernym links from WordNet. \\

\noindent \textbf{Unsatisfiable Clauses:} H has more information than what can be satisfied by T. Example: \\

\noindent 6296 T: A large group of cheerleaders is walking in a parade \\
6296 H: The cheerleaders are parading and wearing black, pink and white uniforms \\
6296 A: NO \\

Coordinated or subordinated clauses in H can be unsatisfiable if T has fewer clauses that H. In the above example, T is composed by a single clause while H has two coordinated clauses and, although the first H's clause can be fully entailed by T, the second one can't be satisfied. Mismatching number of clauses between T and H is detected through the analysis of the sentences' syntactic parse trees.

Any of the above described phenomena is considered enough to reject the entailment, so the TED and DGN models always take into account the output of the Context Analysis module and, in case their conclusions diverge, the decision made on the basis of the contextual information prevails.

\section{Evaluation}\label{sec:eval}
To evaluate our proposed composite syntactic-semantic text entailment approach, we tested the XTE on several datasets, experimenting with varying knowledge bases, built from different lexical resources. The details are given next.

\subsection{Datasets}\label{sec:datasets}
We tested XTE on four datasets:

\noindent\textbf{RTE3 dataset:} the dataset from the third RTE Challenge\footnote{https://www.k4all.org/project/third-recognising-textual-entailment-challenge/} is one of the most traditional and popular text entailment datasets. It contains 1,600 T-H pairs, split into DEV (800 pairs) and TEST (800 pairs) sets, and is balanced, with half positive and half negative examples.

\noindent\textbf{SICK dataset:} SICK\footnote{http://clic.cimec.unitn.it/composes/sick.html} (\textit{Sentences Involving Compositional Knowledge}) is a dataset aimed at the evaluation of compositional distributional semantic models, which, besides the semantic relatedness between sentences, also includes annotations about the entailment relation for the sentence pairs \citep{marelli2014sick}. It is composed of 9,840 pairs, split into TRAIN (4,439 pairs), TRIAL (495 pairs), and TEST (4,906 pairs). Instead of the binary entailment classification, there are three different relations: \textit{entailment}, \textit{contradiction}, and \textit{neutral}. For coherence with the other datasets, we considered both the \textit{contradiction} and \textit{neutral} labels as non-entailment, leading to 29\% positive and 71\% negative examples (the original classification is also unbalanced: around 57\% of the pairs have the label \textit{neutral}).

\noindent\textbf{BPI dataset:} The \textit{Boeing-Princeton-ISI}\footnote{http://www.cs.utexas.edu/users/pclark/bpi-test-suite/} textual entailment test suite was developed specifically to look at entailment problems requiring world knowledge, being syntactically simpler than RTE datasets but more challenging from the semantic viewpoint. It is composed of 250 pairs, 50\% positive and 50\% negative.

\noindent\textbf{GHS dataset:} The \textit{Guardian Headlines Sample}\footnote{https://goo.gl/4iHdbX} is a subset of the Guardian Headlines dataset\footnote{https://goo.gl/XrEwG9}, a set of 32,000 entailment pairs automatically extracted from The Guardian newspaper but not validated. The GHS is a random sample of 800 pairs which have been manually curated, leading to a balanced set of 400 positive and 400 negative examples. It also requires a reasonable amount of world knowledge and is the only dataset fully composed of real-world data, without artificially assembled hypotheses: in positive examples, T is the first sentence of a story and H is its headline, and in negative examples T and H are two random sentences from the same story.

\subsection{Knowledge Bases}\label{sec:kb}
To evaluate the impact of different lexical resources in the entailment results, especially in the justifications generated, we extracted the definitions from four dictionaries: \textbf{WordNet}, the \textbf{Webster's Unabridged Dictionary}\footnote{http://www.gutenberg.org/ebooks/673}, \textbf{Wiktionary}\footnote{https://www.wiktionary.org/}, and the set of definitions extracted from \textbf{Wikipedia} pages provided by \citet{faralli2013java}. Each of the four resulting DKGs differs from the others in some way: The Webster's is an older, conventional dictionary dating from 1913. WordNet and Wiktionary are modern on-line lexicons, but the former is developed by professional lexicographers while the latter is built collaboratively by lay users. Last, Wikipedia is also built collaboratively, but is an encyclopedic, rather than lexical, resource.

The original Webster's dictionary text file was processed so, besides the definitions, the part-of-speech and list of synonyms (when available) for each word could also be extracted\footnote{Extraction in JSON format available at https://github.com/ssvivian/WebstersDictionary}. All the four sets of definitions were submitted to a pre-processing stage intended at filtering potential invalid definitions, including (but not restricted to): verb definitions not beginning with a verb or an adverbial phrase followed by a verb, noun definitions beginning with verbs or prepositions, etc. For the Wikipedia dataset, definitions for named entities were also excluded with the aid of the Stanford Named Entity Recognizer (NER), so the final content could be closer to a regular dictionary. Due to the natural limitations of the NER, many named entity definitions remained in the final set, but this additional filter helped to set a manageable size for the final graph, without leaving out potentially relevant information.

Finally, all the filtered sets of definitions were labeled and converted to an RDF graph, as described in Section \ref{sec:dkg}. Table \ref{tab:kbs} shows the dimensions of each of the resulting graphs.

\begin{table}[ht]
	\centering
	
	\begin{tabular}{lccc}
		\textbf{Resource} & \textbf{Noun Definitions} & \textbf{Verb Definitions} & \textbf{Total} \\ 
		\hline 
		WordNet & 79,939 & 13,760 & 93,699 \\ 
		Webster's & 88,620 & 25,290 & 113,910 \\
		Wiktionary & 390,417 & 73,826 & 464,243 \\ 
		Wikipedia & 859,087 & - & 859,087 \\ 
	\end{tabular}
	\caption{Final dimensions of the definition knowledge graphs used in the interpretable composite text entailment approach}
	\label{tab:kbs}
\end{table}

\subsection{Computing the Thresholds}\label{sec:thresholds}
Two of the most important parameters of the proposed approach are the Tree Edit Distance model's threshold $t$ (Section \ref{sec:ted}), and the Distributional Graph Navigation algorithm's semantic relatedness threshold $\eta$ (Section \ref{sec:dgn}). TED threshold $t$ is computed previously through a training procedure which performs a sequential search to look for the distance that better separates positive examples from the negative ones, and is aimed at maximizing the algorithm's accuracy, in our case, the F-measure score. For training the model, we combined the training portions of the RTE3 and SICK datasets. This combined dataset, herein called RTE+SICK train dataset, compensates for the lack of training data in the other two datasets while still being representative of their syntactic characteristics: the RTE3 is closer in format to the GHS, both having very long text sentences and usually short hypothesis, while the SICK data is more similar to the BPI entries, with both datasets having short to medium-sized text and hypothesis sentences, and usually not a big difference in size between the two sentences composing an entailment pair. After the training is performed over the RTE+SICK dataset, the learned threshold $t$ is used to compute the syntactic entailments for all the four tested datasets (for the RTE3 and SICK datasets, the evaluation is performed on their test portions).

The DGN threshold $\eta$ is computed dynamically so the algorithm can always retrieve the highest scoring entries from a list of candidates. While navigating the knowledge graph, the DGN always retrieves a set of nodes and computes the semantic relatedness $sr$ between each node and the target (see Section \ref{sec:dgn}), which results in a ranked list of scores. Over this list, we perform a \textit{Semantic Differential Analysis}, adapting the method proposed in \citep{freitas2012distributional} to identify score gaps which discriminate between highly semantically related nodes and non-related ones. Given a list of ranked nodes, $S_{0}$ is the score for the node with the maximum relatedness value, $S_{k}$ is the score for the k+1 ranked node and $\delta S_{k,k+1}$ is the semantic differential between two adjacent ranked nodes, that is:

\begin{equation}\label{eq:delta}
\delta S_{k,k+1} = S_{k} - S_{k+1}
\end{equation}

\begin{figure}[ht]
	\centering
	\includegraphics[height=2.5in]{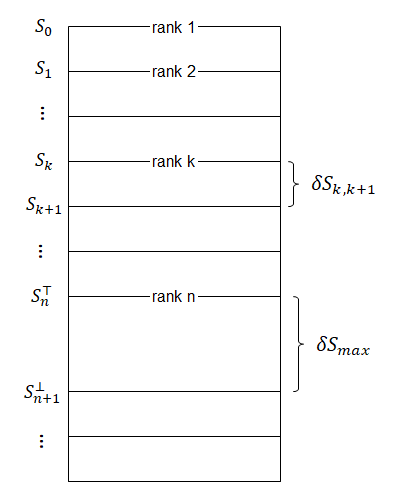}
	\caption{The Semantic Differential Model. $\delta S_{max}$ defines a gap in a ranked list of scores}
	\label{fig:sda}
\end{figure}

The gap in the list, occurring between $S_{n}$ and $S_{n+1}$, is given by $\delta S_{max}$, the maximum semantic differential. $S_{n}$ and $S_{n+1}$ define the top and bottom relatedness values of $\delta S_{max}$, denoted by $S^{\top}_{n}$ and $S^{\bot}_{n+1}$. Figure \ref{fig:sda} illustrates the Semantic Differential Model.

To determine $\eta$ at each step of the graph navigation, we compute $\delta S_{max}$ over the current ranked list of nodes and select the bottom value as the semantic threshold, therefore:

\begin{equation}\label{eq:etavalue}
\eta = S^{\bot}_{n+1}
\end{equation} 

We choose the bottom value, and not the top, which is the one immediately before the gap, in order to keep at least one moderately related node in the list. Since semantic similarity scores depend on the DSM model used to compute them, and, by extension, on the corpus from where the DSM was learned, average (immediately after the gap) scores can have varying meanings when considered in different contexts. By including the bottom relatedness value in the list, we ensure we won't miss potential relevant nodes. If such nodes prove to be irrelevant, the DGN manages to abort their paths at subsequent steps, eliminating any eventual noise.

\subsection{Baselines}\label{sec:baselines}
To show the improvements provided by our composition of entailment methods, we compare our results with three single-technique approaches: a purely syntactic algorithm, a syntactic approach that employs linguistic resources from where shallow semantic information is extracted, and a purely semantic algorithm.

The \textit{Edit Distance} \citep{kouylekov2005recognizing} is the state-of-the-art implementation of the tree edit distance algorithm for recognizing textual entailment, and only considers the syntactic structures of T and H, given by their dependency trees. The \textit{Maximum Entropy Classifier} \citep{wang2008divide} also uses the syntactic dependency trees as features in a classifier which also employs lexical-semantic features from WordNet and VerbOcean (we use the \textit{Base+WN+TP+TPPos+TS\_EN} configuration). Both implementations are provided by the text entailment framework \textit{Excitement Open Platform} (EOP) \citep{magnini2014excitement}, and were also trained on the combined RTE+SICK training dataset (Section \ref{sec:thresholds}). As the semantic-only baseline we use the \textit{Graph Navigation} algorithm using the WordNet definition graph, described in \citep{silva2018recognizing}.

\subsection{Results}\label{sec:results}
Table \ref{tab:results} shows the precision, recall and F-measure (rounded to two decimal places) obtained by each of the XTE configurations, which are given by the DKG used by the Distributional Graph Navigation component, that is, the knowledge bases derived from WordNet (WN), Webster's dictionary (WBT), Wiktionary (WKT), and Wikipedia (WKP), as well as the baselines results.

\begin{table}[ht]
	\resizebox{\textwidth}{!}{
	\begin{tabular}{llll|lll|lll|lll}
		& \multicolumn{3}{c|}{\textbf{RTE3}}                                  & \multicolumn{3}{c|}{\textbf{SICK}}                                  & \multicolumn{3}{c|}{\textbf{BPI}}                                   & \multicolumn{3}{c}{\textbf{GHS}}                                   \\ \cline{2-13} 
		& \textit{\textbf{Pr}} & \textit{\textbf{Re}} & \textit{\textbf{F1}} & \textit{\textbf{Pr}} & \textit{\textbf{Re}} & \textit{\textbf{F1}} & \textit{\textbf{Pr}} & \textit{\textbf{Re}} & \textit{\textbf{F1}} & \textit{\textbf{Pr}} & \textit{\textbf{Re}} & \textit{\textbf{F1}} \\ \cline{2-13} 
		ED         & 0.61                 & 0.51                 & 0.55                 & 0.41                 & 0.76                 & 0.54                 & 0.41                 & 0.45                 & 0.43                 & 0.97                 & 0.15                 & 0.26                 \\
		MEC     & 0.56                 & 0.58                 & 0.57                 & 0.65                 & 0.47                 & 0.55                 & 0.29                 & 0.18                 & 0.23                 & 0.99                 & 0.20                 & 0.33                 \\
		GN      & 0.48                 & 0.32                 & 0.38                 & 0.31                 & 0.35                 & 0.33                 & 0.65                 & 0.54                 & 0.59                 & 0.56                 & 0.50                 & 0.53                 \\ \hline
		XTE (WN)  & 0.57                 & 0.68                 & 0.62                 & 0.50                 & 0.70                 & 0.58                 & 0.56                 & 0.78                 & \textbf{0.65}        & 0.70                 & 0.53                 & \textbf{0.60}        \\
		XTE (WBT) & 0.58                 & 0.65                 & 0.61                 & 0.51                 & 0.69                 & \textbf{0.59}        & 0.53                 & 0.62                 & 0.57                 & 0.69                 & 0.46                 & 0.55                 \\
		XTE (WKT) & 0.59                 & 0.67                 & \textbf{0.63}        & 0.51                 & 0.70                 & \textbf{0.59}        & 0.54                 & 0.70                 & 0.61                 & 0.70                 & 0.45                 & 0.55                 \\
		XTE (WKP) & 0.62      & 0.52                 & 0.56                 & 0.58                 & 0.55                 & 0.57                 & 0.50                 & 0.41                 & 0.45                 & 0.74                     & 0.27                     &  0.40                   
	\end{tabular}}
	\caption{Evaluation results. The upper part shows the baselines, and at the bottom are the proposed composite entailment approach's results. ED = EditDistance, MEC = MaxEntClassifier, GN = GraphNavigation, XTE = ExplainableTextEntailment}
	\label{tab:results}
\end{table}

The first thing that can be noticed is how the results vary across datasets for those baselines that rely on training data, that is, the EditDistance and the MaxEntClassifier algorithms. Both approaches present homogeneous results for the RTE and SICK datasets, for which training data is available, but their accuracy falls significantly for the BPI and GHS datasets. Meanwhile, XTE presents consistent results across all datasets, because we don't rely exclusively on the syntactic information learned at the training phase, but rather balance it with the semantic knowledge extracted from the DKGs. Since these knowledge graphs are commonsense and independent resources, they work homogeneously for any unseen data, making our approach much less training data-dependent. Moreover, for the BPI and GHS datasets, which require more world knowledge, both algorithms show low recall and are surpassed by XTE, adding to the importance of combining and balancing syntactic and semantic information while solving the entailments.

When compared with the semantic-only GraphNavigation approach, XTE also presents much better results, especially for the RTE3 and SICK datasets, which don't have a heavy focus on world knowledge-based entailments. Besides not dealing well with entailments that don't show any semantic relationship (that is, purely syntactic entailments), the GraphNavigation also uses a syntax-based heuristic to define the source-target input pairs and the head words for multi-word expression graph nodes. This heuristic uses part-of-speech tags to find the main components (subject, verb, objects) in a sentence, which may not work well for long, complex sentences, as is the case in the RTE3 and GHS datasets. By using a semantic similarity-based heuristic, we can retrieve better source-target pairs and a higher number of relevant head words, what leads to much better recall and overall F-measure.

\subsubsection{Justifications}\label{sec:justif}
We analyzed the justifications generated for the positive entailments solved by the Distributional Graph Navigation model to assess their correctness and consistency. This evaluation was intended to assess the explanations from a \textit{functional} point of view, that is, to determine if they were accomplishing the task of establishing the right relationship between the right terms in T and H. A deeper, psychological evaluation to assess trustworthiness was out of the scope of this evaluation and is included as future work. That means justifications were evaluated to be ``right'' or ``wrong'' on a high level, but not ``good'' or ``bad'' according to a more subjective user's judgment.

Evaluators were asked to first point the entities establishing a connection between T and H. Then, they should judge if the justification met two requirements: (1) it linked the previously identified entities, and (2) the relationship it describes is the same one intended by the context given by T and H (according to the human judgment).

They then classified justifications into \textit{correct} or \textit{incorrect}. Correct justifications meet both conditions, establishing the right relationship between the relevant entities $e_{1} \in T$ and $e_{2} \in H$, presenting the pertinent information about it and making the reasoning clear. Examples\footnote{Examples from the BPI and RTE3 datasets, respectively.}: \\

\noindent 1.4 T: A council worker cleans up after Tuesday's violence in Budapest. \\
1.4 H: There was damage in Budapest. \\
1.4 A: YES \\[5pt]
Entailment: yes \\
Justification: \\
A \textbf{violence} is a state resulting in injuries and \textbf{destruction} etc. \\
A \textbf{destruction} is a termination of something by causing so much \textbf{damage} to it that it cannot be repaired or no longer exists \\

\noindent 116 T: Vasquez Rocks Natural Area Park is a northern Los Angeles County park acquired by LA County government in the 1970s. \\
116 H: The Vasquez Rocks Natural Area Park is a property of the LA County government. \\
116 A: YES \\[5pt]
Entailment: yes \\
Justification: \\
To \textbf{acquire} is to come into the \textbf{possession} \\
A \textbf{possession} is an act of having and controlling \textbf{property} \\

Incorrect justifications do not meet one or both aforementioned conditions, establishing a relationship between the wrong pair of entities, that is, entities that, although being semantically related, do not establish a logical link between T and H, or being too vague, linking the correct pair of entities $e_{1} \in T$ and $e_{2} \in H$ but giving only superficial and insufficient information about their semantic relationship. Examples\footnote{Examples from the RTE3 dataset.}: \\

\noindent 149 T: Joining Pinkerton at the Chamber of Commerce, is Lindsey Beverly, who will be the new executive assistant. \\
149 H: Pinkerton works with Beverly. \\
149 A: YES \\[5pt]
Entailment: yes \\
Justification: \\
An \textbf{assistant} is a person who contributes to the fulfillment of a need or furtherance of an \textbf{effort} or purpose \\
\textbf{Effort} is synonym of \textbf{work} \\

\noindent 110 T: Leloir was promptly given the Premio de la Sociedad Cient\'{\i}fica Argentina, one of few to receive such a prize in a country in which he was a foreigner. \\
110 H: Leloir won the Premio de la Sociedad Cient\'{\i}fica Argentina. \\
110 A: YES \\[5pt]
Entailment: yes \\
Justification: \\
To \textbf{receive} is a way of to \textbf{take} \\
To \textbf{take} is synonym of to \textbf{win} \\

In the first example, even though the semantic relationship between ``assistant'' and ``work'' may seem consistent, the expressions that establish the entailment relation are ``join'' and ``work with''. In the second case, although the justification links the correct entities establishing the entailment, that is, ``receive'' and ``win'', the explanation, made through the verb ``take'' with no complements, sounds vague and not informative enough. Definitions for verbs tend to be less detailed than those for nouns, and many times expressed in terms of very broad supertypes (like ``take'' in the second example), leading justifications generated from paths containing verb entity nodes to be more prone to vagueness.

The distribution of correct and incorrect justifications is given in Table \ref{tab:justifdist}. The evaluation was performed over the results obtained with the WordNet graph, which is the knowledge base that yields the overall best results across datasets. A more detailed comparison of all the tested DKGs is given in the Section \ref{sec:comparison}.

\begin{table}[ht]
	\centering
	\begin{tabular}{lcc}
		\textbf{Dataset} & \textbf{Correct Justifications} & \textbf{Incorrect Justifications} \\ \hline
		RTE3             &  50.3\%                         &  49.7\%                           \\
		SICK             &  77.1\%                         &  22.9\%                           \\
		BPI              &  61.3\%                         &  38.7\%                           \\
		GHS              &  43.3\%                         &  56.7\%                                
	\end{tabular}
	\caption{Distribution of correct and incorrect justifications}
	\label{tab:justifdist}
\end{table}

As can be seen in Table \ref{tab:justifdist}, the distribution of correct and incorrect justifications varies depending on the dataset, with SICK and BPI showing the best results. In common, these two datasets have relatively short text and hypothesis sentences, which favors the correct identification of source-target word pairs. On the other hand, the RTE3 and GHS datasets have short hypothesis sentences but very long text sentences and the larger the sentence, the bigger the number of possible source-target pairs, so the likelihood of finding a relationship that doesn't necessarily leads to the entailment increases. The GHS is a particularly challenging dataset because the text T corresponds to the the first sentence of a news story, usually expanding the idea highly condensed in the headline, that is, the hypothesis H. The two sentences are very semantically related as a whole, so many concepts in T will have some strong relationship with (sometimes the same) concepts in H, leading to many explanations where, although the relationship by itself may be right, it is not the most suitable answer for the entailment decision, as already shown in the example 149 above, hence the higher number of incorrect justifications.

Although there is still much room for improvement, our approach for generating natural language justifications proved to be a viable solution especially in view of the fact that it employs an unsupervised technique and relies on already existing knowledge sources, yielding reasonable results without the costs and training data-dependency of supervised methods. 

\subsubsection{Comparing Definition Knowledge Graphs}\label{sec:comparison}
Besides the improvements in the quantitative results, the interpretable characteristic of XTE represents an important contribution: providing human-like explanations for the entailment decisions whenever a more complex semantic relationship is involved translates into a concrete gain for the final user, who can understand and judge the system's rationale. The justifications, though, depend heavily on the graph knowledge base employed in the entailment recognition. Overall, WordNet, Webster's and Wiktionary graphs deliver close results for the RTE3 and SICK datasets, but WordNet stands out for the more world knowledge-demanding BPI and GHS datasets, as can be seen in Table \ref{tab:results}.

The impact of each DKG can be better measured by the \textit{recall} obtained when they are queried: the more useful information the graph contains, the more paths (meaning semantic relationships between source and target words) can be found and, consequently, more entailments can be recognized. Again, WordNet, Webster's and Wiktionary graphs show comparable recalls for the RTE3 and SICK datasets, but WordNet presents a much better recall for BPI and GHS. The Wikipedia graph, on the other hand, shows lower recall for all of the datasets, especially for the more knowledge-oriented BPI and GHS, despite being the larger knowledge base. This happens because Wikipedia, besides not defining verbs, privileges the definitions of people, places, arts and entertainment artifacts (films, books, songs, etc.) and other entities expressed by proper nouns. In fact, Wikipedia lacks definitions for many concepts present in the datasets for which relationships are sought: ``violence'' and ``damage'', ``signatory'' and ``agreement'', ``decontamination'' and ``contaminants'', or ``bet'' and ``gamble'', to name a few. This shows that, for the entailment task, the content type is more relevant than the amount of information in the graph. The WordNet graph, for example, corresponds to roughly only 10\% of the Wikipedia graph, but contains far more common nouns denoting basic language concepts, better matching the task requirements.

The Wiktionary graph has a good coverage of common nouns, comparable to WordNet, but in some cases the completeness of its definitions may represent an issue: if not enough information is contained in the definition, that is, the definition of an entity fails to mention other entities it is essentially related to, paths will start but won't reach the target. Built by expert lexicographers, the WordNet definitions tend to follow some patterns and are more prone to cover essential attributes. On the other hand, in a collaborative environment, despite the larger volume of information that can be generated, high-quality standards can't always be ensured. This is the reason why, in spite of its much larger dimensions, the Wiktionary graph can't always surpass the WordNet one, yielding lower recall for the BPI and GHS datasets. Again, the coverage and regularity of the knowledge base contents prove to be more important than its size.

As for the Webster's graph, what was observed as an issue is the oldness of its source: dating from 1913, the Webster's Unabridged Dictionary naturally also covers all the most common language concepts, but, besides sometimes registering obsolete forms, like ``camera obscura'' for [photographic] ``camera'', lacks many modern concepts or concepts that were not of widespread use back then, such as ``WMD'' (Weapons of Mass Destruction), ``website'', ``terrorist act'' or ``recall'' (in the sense of ``defective products callback''). Such modern concepts are frequent in press content, so datasets like the GHS, totally generated from newspaper content, and the BPI, also derived from news content, may pose a challenge to this knowledge graph, which is confirmed by the lower recall (compared to WordNet) returned for these two datasets when the Webster's graph is used.

The justifications generated by each of the graphs are comparable in quality, with WordNet and Wikipedia graphs offering slightly more detailed explanations. An example from the GHS dataset, explained by the WordNet graph: \\

\noindent 18623 T: GlaxoSmithKline has been forced to set aside 220m to settle anti-trust cases in the US over its anti-inflammatory drug Relafen. \\
18623 H: Glaxo hit by 220m US court blow \\
18623 A: YES \\[5pt]
Entailment: yes \\
Justification: \\
A \textbf{case} is a term for any proceeding in a \textbf{court of law} whereby an individual seeks a legal remedy \\
\textbf{Court of law} is synonym of \textbf{court} \\

\noindent From the RTE3 dataset, explained by the Webster's graph: \\

\noindent 158 T: Mr. Gotti, who is already serving nine years on extortion charges, was sentenced to an additional 25 years by Judge Richard D. Casey of Federal District Court. \\
158 H: Gotti was accused of extortion. \\
158 A: YES \\[5pt]
Entailment: yes \\
Justification: \\
A \textbf{charge} is a kind of \textbf{accusation} \\
An \textbf{accusation} is an act of \textbf{accusing} or charging with a crime or with a lighter offense \\

\noindent From the SICK dataset, explained by the Wiktionary graph: \\

\noindent 1459 T: A man is exercising \\
1459 H: A man is doing physical activity \\
1459 A: YES \\[5pt]
Entailment: yes \\
Justification: \\
To \textbf{exercise} is to perform \textbf{physical activity} for health or training \\

\noindent From the BPI dataset, explained by the Wikipedia graph: \\

\noindent 17.4 T: A Union Pacific freight train hit five people. \\
17.4 H: The train was moving along a railroad track. \\
17.4 A: YES \\[5pt]
Entailment: yes \\
Justification: \\
A \textbf{freight train} is a group of freight cars or goods wagons hauled by one or more locomotives on a \textbf{railway} [...] \\
\textbf{Railway} is synonym of \textbf{railroad track} \\

In summary, the WordNet graph presents the best recall across all datasets, due to its good term coverage and definitions' completeness. The Wiktionary and Webster's graphs also show good recall but their performance can be weakened by the incompleteness resulting from the amateur nature of the definitions creation process, in the Wiktionary case, or by the lack of modern terms which are frequent in contemporary language, in the Webster's instance. The Wikipedia graph, due to its encyclopedic nature, has well constructed and complete definitions, but lacks many basic concepts, yielding the lowest recall regardless of the dataset.

\subsection{Limitations}
Despite outperforming state-of-the-art text entailment algorithms, the proposed approach still does not present very high accuracy, especially if compared with accuracy-driven modern NLI models. In fact, solving knowledge-demanding semantic entailments in an interpretable way is a complex task with still much room for new developments. Analyzing our approach results to understand the cause of false positive and false negatives, we identified some limitations that can be better addressed in future work.

Wrong entailment decisions can be caused by both internal and external factors. Internal factors indicate the current limitations of the proposed approach, while external factors refer to third-party resources for which no fine-tuning is possible. Undetected context information and wrong semantic relationship assumption are the most common internal sources of errors. External sources include syntactic parser error and knowledge base incompleteness. Knowledge base-related issues affect not only the system accuracy but also the quality of the resulting justifications.

We observed that the errors caused by internal factors occur much more often when there is a \textit{neutral} relationship between T and H, that is, there is no entailment but also no contradiction. In such cases, more obvious context information is not always available, indicating that our approach needs to adopt further analyses and resources to better deal with these scenarios. Regarding external factors, improving knowledge base coverage could also bring considerable gains, both quantitatively and qualitatively.

\section{Conclusion}\label{sec:conc}
Recognizing textual entailment is an important task whose outputs feeds many other NLP applications. It can involve a wide range of different linguistic and semantic phenomena, and identifying such phenomena and using the most suitable techniques for each of them is key to better accuracy. We presented XTE, an interpretable composite approach for recognizing text entailment that, employing a routing mechanism that analyzes the overlap between the text and hypothesis, decides whether entailment pairs should be dealt with syntactically or semantically. For those pairs predominantly showing structural -- that is, syntactic -- differences, we use a Relative Tree Edit Distance model over dependency parse trees, and for pairs where a semantic relationship exists we employ a Distributional Graph Navigation model over knowledge bases composed of structured dictionary definitions. 

Whenever the entailment is solved semantically, the paths found in the graph KBs by the DGN model are used to render the entailment interpretable, providing natural language, human-like justifications for the entailment decision. By explaining the system's reasoning steps, we make it possible to interpret and understand its underlying inference model, taking the entailment decision out of the numerical score black box.

We built definition knowledge graphs from four different lexical resources: WordNet, the Webster's Unabridged Dictionary, Wiktionary and Wikipedia, and assessed how each of them impacted our approach results and influenced its interpretability. Given that text entailment deals with language variability, we observed that knowledge graphs covering the most basic, everyday language concepts yield the best results, so regular dictionaries, such as WordNet, Webster's and Wiktionary are more useful than encyclopedic KBs like Wikipedia for this task. We also found that definitions created by lexicographers under a controlled environment tend to be more complete and, consequently, provide better recall and somewhat more detailed justifications than those created in collaborative environments by lay users. Furthermore, contemporary resources can show some advantage over older dictionaries for containing modern terms frequently occurring in the present-day language but absent from ancient lexicons like Webster's dictionary. 

Our interpretable composite approach outperforms entailment algorithms that employ a single technique, be it syntactic or semantic, to tackle all types of entailments, and is less dependent on training data, since it combines learned parameters with independent, external commonsense knowledge which works well regardless of the dataset. By testing and comparing several graph knowledge bases, we show that the use of external world knowledge not only improve quantitative results, but is also a valuable feature for increasing intelligent systems' interpretability. Recent advancements, such as the new NLI techniques, which presents good results but are mostly accuracy-driven, could benefit from these resources to also explain themselves, staying in line with Explainable AI requirements.

As future work, especially to overcome the current limitations, we believe that the system performance can be improved by the exploration of more advanced graph analysis techniques to enable the aggregation of multiple knowledge bases for better information coverage. The use of further resources and features for capturing more complex or subtler context information can also improve the overall accuracy. Finally, a complementary qualitative evaluation of the justifications for assessing their trustworthiness from the user's perspective, through a more sophisticated psychological study, can contribute to reinforcing the usefulness of the system's explainable dimension.

\bibliographystyle{elsarticle-num-names}
\bibliography{XTE_VSS_et_al}

\begin{thebibliography}{62}
\expandafter\ifx\csname natexlab\endcsname\relax\def\natexlab#1{#1}\fi
\providecommand{\url}[1]{\texttt{#1}}
\providecommand{\href}[2]{#2}
\providecommand{\path}[1]{#1}
\providecommand{\DOIprefix}{doi:}
\providecommand{\ArXivprefix}{arXiv:}
\providecommand{\URLprefix}{URL: }
\providecommand{\Pubmedprefix}{pmid:}
\providecommand{\doi}[1]{\href{http://dx.doi.org/#1}{\path{#1}}}
\providecommand{\Pubmed}[1]{\href{pmid:#1}{\path{#1}}}
\providecommand{\bibinfo}[2]{#2}
\ifx\xfnm\relax \def\xfnm[#1]{\unskip,\space#1}\fi
\bibitem[{Dagan et~al.(2013)Dagan, Roth, Sammons, and
  Zanzotto}]{dagan2013recognizing}
\bibinfo{author}{I.~Dagan}, \bibinfo{author}{D.~Roth},
  \bibinfo{author}{M.~Sammons}, \bibinfo{author}{F.~M. Zanzotto},
\newblock \bibinfo{title}{Recognizing textual entailment: Models and
  applications},
\newblock \bibinfo{journal}{Synthesis Lectures on Human Language Technologies}
  \bibinfo{volume}{6} (\bibinfo{year}{2013}) \bibinfo{pages}{1--220}.
\bibitem[{Dagan et~al.(2006)Dagan, Glickman, and Magnini}]{dagan2006pascal}
\bibinfo{author}{I.~Dagan}, \bibinfo{author}{O.~Glickman},
  \bibinfo{author}{B.~Magnini},
\newblock \bibinfo{title}{The {PASCAL} recognising textual entailment
  challenge},
\newblock in: \bibinfo{booktitle}{Machine learning challenges: evaluating
  predictive uncertainty, visual object classification, and recognising textual
  entailment}, \bibinfo{publisher}{Springer}, \bibinfo{year}{2006}, pp.
  \bibinfo{pages}{177--190}.
\bibitem[{Gunning(2017)}]{gunning2017explainable}
\bibinfo{author}{D.~Gunning},
\newblock \bibinfo{title}{Explainable artificial intelligence ({XAI})},
\newblock \bibinfo{journal}{Defense Advanced Research Projects Agency
  ({DARPA})}  (\bibinfo{year}{2017}).
\bibitem[{Ghuge and Bhattacharya(2014)}]{ghuge2014survey}
\bibinfo{author}{S.~Ghuge}, \bibinfo{author}{A.~Bhattacharya},
\newblock \bibinfo{title}{Survey in textual entailment},
\newblock \bibinfo{journal}{Center for Indian Language Technology}
  (\bibinfo{year}{2014}).
\bibitem[{Glickman and Dagan(2005)}]{glickman2005probabilistic}
\bibinfo{author}{O.~Glickman}, \bibinfo{author}{I.~Dagan},
\newblock \bibinfo{title}{A probabilistic setting and lexical cooccurrence
  model for textual entailment},
\newblock in: \bibinfo{booktitle}{Proceedings of the {ACL} Workshop on
  Empirical Modeling of Semantic Equivalence and Entailment},
  \bibinfo{organization}{Association for Computational Linguistics},
  \bibinfo{year}{2005}, pp. \bibinfo{pages}{43--48}.
\bibitem[{P{\'e}rez and Alfonseca(2005)}]{perez2005application}
\bibinfo{author}{D.~P{\'e}rez}, \bibinfo{author}{E.~Alfonseca},
\newblock \bibinfo{title}{Application of the {B}leu algorithm for recognising
  textual entailments},
\newblock in: \bibinfo{booktitle}{Proceedings of the First Challenge Workshop
  Recognising Textual Entailment}, \bibinfo{year}{2005}, pp.
  \bibinfo{pages}{9--12}.
\bibitem[{Newman et~al.(2005)Newman, Stokes, Dunnion, and
  Carthy}]{newman2005ucd}
\bibinfo{author}{E.~Newman}, \bibinfo{author}{N.~Stokes},
  \bibinfo{author}{J.~Dunnion}, \bibinfo{author}{J.~Carthy},
\newblock \bibinfo{title}{{UCD} {IIRG} approach to the textual entailment
  challenge},
\newblock in: \bibinfo{booktitle}{Proceedings of the PASCAL Challenges Workshop
  on Recognising Textual Entailment}, \bibinfo{year}{2005}, pp.
  \bibinfo{pages}{53--56}.
\bibitem[{MacCartney et~al.(2008)MacCartney, Galley, and
  Manning}]{maccartney2008phrase}
\bibinfo{author}{B.~MacCartney}, \bibinfo{author}{M.~Galley},
  \bibinfo{author}{C.~D. Manning},
\newblock \bibinfo{title}{A phrase-based alignment model for natural language
  inference},
\newblock in: \bibinfo{booktitle}{Proceedings of the conference on empirical
  methods in natural language processing}, \bibinfo{organization}{Association
  for Computational Linguistics}, \bibinfo{year}{2008}, pp.
  \bibinfo{pages}{802--811}.
\bibitem[{Wang and Neumann(2008)}]{wang2008accuracy}
\bibinfo{author}{R.~Wang}, \bibinfo{author}{G.~Neumann},
\newblock \bibinfo{title}{An accuracy-oriented divide-and-conquer strategy for
  recognizing textual entailment},
\newblock in: \bibinfo{booktitle}{Proceedings of the Text Analysis Conference
  {TAC}}, \bibinfo{year}{2008}.
\bibitem[{Sammons et~al.(2009)Sammons, Vydiswaran, Vieira, Johri, Chang,
  Goldwasser, Srikumar, Kundu, Tu, Small, Rule, Do, and
  Roth}]{sammons2009relation}
\bibinfo{author}{M.~Sammons}, \bibinfo{author}{V.~V. Vydiswaran},
  \bibinfo{author}{T.~Vieira}, \bibinfo{author}{N.~Johri},
  \bibinfo{author}{M.-W. Chang}, \bibinfo{author}{D.~Goldwasser},
  \bibinfo{author}{V.~Srikumar}, \bibinfo{author}{G.~Kundu},
  \bibinfo{author}{Y.~Tu}, \bibinfo{author}{K.~Small},
  \bibinfo{author}{J.~Rule}, \bibinfo{author}{Q.~Do},
  \bibinfo{author}{D.~Roth},
\newblock \bibinfo{title}{Relation alignment for textual entailment
  recognition},
\newblock in: \bibinfo{booktitle}{Proceedings of the Text Analysis Conference
  {TAC}}, \bibinfo{year}{2009}.
\bibitem[{Harmeling(2009)}]{harmeling2009inferring}
\bibinfo{author}{S.~Harmeling},
\newblock \bibinfo{title}{Inferring textual entailment with a probabilistically
  sound calculus},
\newblock \bibinfo{journal}{Natural Language Engineering} \bibinfo{volume}{15}
  (\bibinfo{year}{2009}) \bibinfo{pages}{459--477}.
\bibitem[{Stern and Dagan(2011)}]{stern2011confidence}
\bibinfo{author}{A.~Stern}, \bibinfo{author}{I.~Dagan},
\newblock \bibinfo{title}{A confidence model for syntactically-motivated
  entailment proofs},
\newblock in: \bibinfo{booktitle}{Proceedings of the International Conference
  Recent Advances in Natural Language Processing 2011}, \bibinfo{year}{2011},
  pp. \bibinfo{pages}{455--462}.
\bibitem[{Kouylekov and Magnini(2005)}]{kouylekov2005recognizing}
\bibinfo{author}{M.~Kouylekov}, \bibinfo{author}{B.~Magnini},
\newblock \bibinfo{title}{Recognizing textual entailment with tree edit
  distance algorithms},
\newblock in: \bibinfo{booktitle}{Proceedings of the First Challenge Workshop
  Recognising Textual Entailment}, \bibinfo{year}{2005}, pp.
  \bibinfo{pages}{17--20}.
\bibitem[{Wang and Neumann(2008)}]{wang2008divide}
\bibinfo{author}{R.~Wang}, \bibinfo{author}{G.~Neumann},
\newblock \bibinfo{title}{An divide-and-conquer strategy for recognizing
  textual entailment},
\newblock in: \bibinfo{booktitle}{Proceedings of the Text Analysis Conference,
  Gaithersburg, MD}, \bibinfo{year}{2008}.
\bibitem[{Zanoli and Colombo(2016)}]{zanoli2016transformation}
\bibinfo{author}{R.~Zanoli}, \bibinfo{author}{S.~Colombo},
\newblock \bibinfo{title}{A transformation-driven approach for recognizing
  textual entailment},
\newblock \bibinfo{journal}{Natural Language Engineering}
  (\bibinfo{year}{2016}) \bibinfo{pages}{1--28}.
\bibitem[{Jimenez et~al.(2014)Jimenez, Due\~{n}as, Baquero, and
  Gelbukh}]{jimenez2014unal}
\bibinfo{author}{S.~Jimenez}, \bibinfo{author}{G.~Due\~{n}as},
  \bibinfo{author}{J.~Baquero}, \bibinfo{author}{A.~Gelbukh},
\newblock \bibinfo{title}{{UNAL-NLP}: Combining soft cardinality features for
  semantic textual similarity, relatedness and entailment},
\newblock in: \bibinfo{booktitle}{Proceedings of the 8th International Workshop
  on Semantic Evaluation (SemEval 2014)}, \bibinfo{year}{2014}, pp.
  \bibinfo{pages}{732--742}.
\bibitem[{Zhao et~al.(2014)Zhao, Zhu, and Lan}]{zhao2014ecnu}
\bibinfo{author}{J.~Zhao}, \bibinfo{author}{T.~Zhu}, \bibinfo{author}{M.~Lan},
\newblock \bibinfo{title}{{ECNU}: One stone two birds: Ensemble of heterogenous
  measures for semantic relatedness and textual entailment},
\newblock in: \bibinfo{booktitle}{Proceedings of the 8th International Workshop
  on Semantic Evaluation (SemEval 2014)}, \bibinfo{year}{2014}, pp.
  \bibinfo{pages}{271--277}.
\bibitem[{Zhang et~al.(2017)Zhang, Chen, Liu, Liu, and Lv}]{zhang2017context}
\bibinfo{author}{K.~Zhang}, \bibinfo{author}{E.~Chen},
  \bibinfo{author}{Q.~Liu}, \bibinfo{author}{C.~Liu}, \bibinfo{author}{G.~Lv},
\newblock \bibinfo{title}{A context-enriched neural network method for
  recognizing lexical entailment},
\newblock in: \bibinfo{booktitle}{{AAAI}}, \bibinfo{year}{2017}, pp.
  \bibinfo{pages}{3127--3134}.
\bibitem[{Silva et~al.(2018)Silva, Freitas, and
  Handschuh}]{silva2018recognizing}
\bibinfo{author}{V.~S. Silva}, \bibinfo{author}{A.~Freitas},
  \bibinfo{author}{S.~Handschuh},
\newblock \bibinfo{title}{Recognizing and justifying text entailment through
  distributional navigation on definition graphs},
\newblock in: \bibinfo{booktitle}{{AAAI}}, \bibinfo{year}{2018}.
\bibitem[{Clark and Harrison(2009)}]{clark2009inference}
\bibinfo{author}{P.~Clark}, \bibinfo{author}{P.~Harrison},
\newblock \bibinfo{title}{An inference-based approach to recognizing
  entailment},
\newblock in: \bibinfo{booktitle}{{TAC}}, \bibinfo{year}{2009}.
\bibitem[{Raina et~al.(2005)Raina, Ng, and Manning}]{raina2005robust}
\bibinfo{author}{R.~Raina}, \bibinfo{author}{A.~Y. Ng}, \bibinfo{author}{C.~D.
  Manning},
\newblock \bibinfo{title}{Robust textual inference via learning and abductive
  reasoning},
\newblock in: \bibinfo{booktitle}{{AAAI}}, \bibinfo{year}{2005}, pp.
  \bibinfo{pages}{1099--1105}.
\bibitem[{Voorhees(2008)}]{voorhees2008contradictions}
\bibinfo{author}{E.~M. Voorhees},
\newblock \bibinfo{title}{Contradictions and justifications: Extensions to the
  textual entailment task},
\newblock in: \bibinfo{booktitle}{ACL}, \bibinfo{year}{2008}, pp.
  \bibinfo{pages}{63--71}.
\bibitem[{Frisse(1988)}]{frisse1988searching}
\bibinfo{author}{M.~E. Frisse},
\newblock \bibinfo{title}{Searching for information in a hypertext medical
  handbook},
\newblock \bibinfo{journal}{Communications of the {ACM}} \bibinfo{volume}{31}
  (\bibinfo{year}{1988}) \bibinfo{pages}{880--886}.
\bibitem[{Gudivada et~al.(1997)Gudivada, Raghavan, Grosky, and
  Kasanagottu}]{gudivada1997information}
\bibinfo{author}{V.~N. Gudivada}, \bibinfo{author}{V.~V. Raghavan},
  \bibinfo{author}{W.~I. Grosky}, \bibinfo{author}{R.~Kasanagottu},
\newblock \bibinfo{title}{Information retrieval on the world wide web},
\newblock \bibinfo{journal}{{IEEE} Internet Computing} \bibinfo{volume}{1}
  (\bibinfo{year}{1997}) \bibinfo{pages}{58--68}.
\bibitem[{Aggarwal and Wang(2011)}]{aggarwal2011text}
\bibinfo{author}{C.~C. Aggarwal}, \bibinfo{author}{H.~Wang},
\newblock \bibinfo{title}{Text mining in social networks},
\newblock in: \bibinfo{booktitle}{Social network data analytics},
  \bibinfo{publisher}{Springer}, \bibinfo{year}{2011}, pp.
  \bibinfo{pages}{353--378}.
\bibitem[{Ganesan et~al.(2010)Ganesan, Zhai, and Han}]{ganesan2010opinosis}
\bibinfo{author}{K.~Ganesan}, \bibinfo{author}{C.~Zhai},
  \bibinfo{author}{J.~Han},
\newblock \bibinfo{title}{Opinosis: a graph-based approach to abstractive
  summarization of highly redundant opinions},
\newblock in: \bibinfo{booktitle}{Proceedings of the 23rd international
  conference on computational linguistics}, \bibinfo{organization}{Association
  for Computational Linguistics}, \bibinfo{year}{2010}, pp.
  \bibinfo{pages}{340--348}.
\bibitem[{Paul et~al.(2016)Paul, Rettinger, Mogadala, Knoblock, and
  Szekely}]{paul2016efficient}
\bibinfo{author}{C.~Paul}, \bibinfo{author}{A.~Rettinger},
  \bibinfo{author}{A.~Mogadala}, \bibinfo{author}{C.~A. Knoblock},
  \bibinfo{author}{P.~Szekely},
\newblock \bibinfo{title}{Efficient graph-based document similarity},
\newblock in: \bibinfo{booktitle}{International Semantic Web Conference},
  \bibinfo{organization}{Springer}, \bibinfo{year}{2016}, pp.
  \bibinfo{pages}{334--349}.
\bibitem[{Kotlerman et~al.(2015)Kotlerman, Dagan, Magnini, and
  Bentivogli}]{kotlerman2015textual}
\bibinfo{author}{L.~Kotlerman}, \bibinfo{author}{I.~Dagan},
  \bibinfo{author}{B.~Magnini}, \bibinfo{author}{L.~Bentivogli},
\newblock \bibinfo{title}{Textual entailment graphs},
\newblock \bibinfo{journal}{Natural Language Engineering} \bibinfo{volume}{21}
  (\bibinfo{year}{2015}) \bibinfo{pages}{699--724}.
\bibitem[{Bowman et~al.(2015)Bowman, Angeli, Potts, and
  Manning}]{bowman2015large}
\bibinfo{author}{S.~R. Bowman}, \bibinfo{author}{G.~Angeli},
  \bibinfo{author}{C.~Potts}, \bibinfo{author}{C.~D. Manning},
\newblock \bibinfo{title}{A large annotated corpus for learning natural
  language inference},
\newblock in: \bibinfo{booktitle}{Conference on Empirical Methods in Natural
  Language Processing ({EMNLP})}, \bibinfo{organization}{Association for
  Computational Linguistics ({ACL})}, \bibinfo{year}{2015}.
\bibitem[{Williams et~al.(2018)Williams, Nangia, and
  Bowman}]{williams2018broad}
\bibinfo{author}{A.~Williams}, \bibinfo{author}{N.~Nangia},
  \bibinfo{author}{S.~Bowman},
\newblock \bibinfo{title}{A broad-coverage challenge corpus for sentence
  understanding through inference},
\newblock in: \bibinfo{booktitle}{Proceedings of the 2018 Conference of the
  North American Chapter of the Association for Computational Linguistics:
  Human Language Technologies, Volume 1 (Long Papers)},
  volume~\bibinfo{volume}{1}, \bibinfo{year}{2018}, pp.
  \bibinfo{pages}{1112--1122}.
\bibitem[{Rockt{\"a}schel et~al.(2016)Rockt{\"a}schel, Grefenstette, Hermann,
  Ko{\v{c}}isk{\`y}, and Blunsom}]{rocktaschel2016reasoning}
\bibinfo{author}{T.~Rockt{\"a}schel}, \bibinfo{author}{E.~Grefenstette},
  \bibinfo{author}{K.~M. Hermann}, \bibinfo{author}{T.~Ko{\v{c}}isk{\`y}},
  \bibinfo{author}{P.~Blunsom},
\newblock \bibinfo{title}{Reasoning about entailment with neural attention},
\newblock in: \bibinfo{booktitle}{Proceedings of the International Conference
  on Learning Representations}, \bibinfo{year}{2016}.
\bibitem[{Wang and Jiang(2016)}]{wang2016learning}
\bibinfo{author}{S.~Wang}, \bibinfo{author}{J.~Jiang},
\newblock \bibinfo{title}{Learning natural language inference with {LSTM}},
\newblock in: \bibinfo{booktitle}{Proceedings of the 2016 Conference of the
  North American Chapter of the Association for Computational Linguistics:
  Human Language Technologies}, \bibinfo{year}{2016}, pp.
  \bibinfo{pages}{1442--1451}.
\bibitem[{Chen et~al.(2017)Chen, Zhu, Ling, Wei, Jiang, and
  Inkpen}]{chen2017enhanced}
\bibinfo{author}{Q.~Chen}, \bibinfo{author}{X.~Zhu}, \bibinfo{author}{Z.-H.
  Ling}, \bibinfo{author}{S.~Wei}, \bibinfo{author}{H.~Jiang},
  \bibinfo{author}{D.~Inkpen},
\newblock \bibinfo{title}{Enhanced {LSTM} for natural language inference},
\newblock in: \bibinfo{booktitle}{Proceedings of the 55th Annual Meeting of the
  Association for Computational Linguistics (Volume 1: Long Papers)},
  \bibinfo{year}{2017}, pp. \bibinfo{pages}{1657--1668}.
\bibitem[{Parikh et~al.(2016)Parikh, T{\"a}ckstr{\"o}m, Das, and
  Uszkoreit}]{parikh2016decomposable}
\bibinfo{author}{A.~Parikh}, \bibinfo{author}{O.~T{\"a}ckstr{\"o}m},
  \bibinfo{author}{D.~Das}, \bibinfo{author}{J.~Uszkoreit},
\newblock \bibinfo{title}{A decomposable attention model for natural language
  inference},
\newblock in: \bibinfo{booktitle}{Proceedings of the 2016 Conference on
  Empirical Methods in Natural Language Processing}, \bibinfo{year}{2016}, pp.
  \bibinfo{pages}{2249--2255}.
\bibitem[{Liu et~al.(2016)Liu, Sun, Lin, and Wang}]{liu2016learning}
\bibinfo{author}{Y.~Liu}, \bibinfo{author}{C.~Sun}, \bibinfo{author}{L.~Lin},
  \bibinfo{author}{X.~Wang},
\newblock \bibinfo{title}{Learning natural language inference using
  bidirectional {LSTM} model and inner-attention},
\newblock \bibinfo{journal}{arXiv preprint arXiv:1605.09090}
  (\bibinfo{year}{2016}).
\bibitem[{Im and Cho(2017)}]{im2017distance}
\bibinfo{author}{J.~Im}, \bibinfo{author}{S.~Cho},
\newblock \bibinfo{title}{Distance-based self-attention network for natural
  language inference},
\newblock \bibinfo{journal}{arXiv preprint arXiv:1712.02047}
  (\bibinfo{year}{2017}).
\bibitem[{Gong et~al.(2017)Gong, Luo, and Zhang}]{gong2017natural}
\bibinfo{author}{Y.~Gong}, \bibinfo{author}{H.~Luo},
  \bibinfo{author}{J.~Zhang},
\newblock \bibinfo{title}{Natural language inference over interaction space},
\newblock \bibinfo{journal}{arXiv preprint arXiv:1709.04348}
  (\bibinfo{year}{2017}).
\bibitem[{Kim et~al.(2019)Kim, Kang, and Kwak}]{kim2019semantic}
\bibinfo{author}{S.~Kim}, \bibinfo{author}{I.~Kang}, \bibinfo{author}{N.~Kwak},
\newblock \bibinfo{title}{Semantic sentence matching with densely-connected
  recurrent and co-attentive information},
\newblock in: \bibinfo{booktitle}{Proceedings of the AAAI Conference on
  Artificial Intelligence}, volume~\bibinfo{volume}{33}, \bibinfo{year}{2019},
  pp. \bibinfo{pages}{6586--6593}.
\bibitem[{Chen et~al.(2018)Chen, Zhu, Ling, Inkpen, and Wei}]{chen2018neural}
\bibinfo{author}{Q.~Chen}, \bibinfo{author}{X.~Zhu}, \bibinfo{author}{Z.-H.
  Ling}, \bibinfo{author}{D.~Inkpen}, \bibinfo{author}{S.~Wei},
\newblock \bibinfo{title}{Neural natural language inference models enhanced
  with external knowledge},
\newblock in: \bibinfo{booktitle}{Proceedings of the 56th Annual Meeting of the
  Association for Computational Linguistics (Volume 1: Long Papers)},
  \bibinfo{year}{2018}, pp. \bibinfo{pages}{2406--2417}.
\bibitem[{Wang et~al.(2019)Wang, Kapanipathi, Musa, Yu, Talamadupula,
  Abdelaziz, Chang, Fokoue, Makni, Mattei, and Witbrock}]{wang2019improving}
\bibinfo{author}{X.~Wang}, \bibinfo{author}{P.~Kapanipathi},
  \bibinfo{author}{R.~Musa}, \bibinfo{author}{M.~Yu},
  \bibinfo{author}{K.~Talamadupula}, \bibinfo{author}{I.~Abdelaziz},
  \bibinfo{author}{M.~Chang}, \bibinfo{author}{A.~Fokoue},
  \bibinfo{author}{B.~Makni}, \bibinfo{author}{N.~Mattei},
  \bibinfo{author}{M.~Witbrock},
\newblock \bibinfo{title}{Improving natural language inference using external
  knowledge in the science questions domain},
\newblock in: \bibinfo{booktitle}{Proceedings of the {AAAI} Conference on
  Artificial Intelligence}, volume~\bibinfo{volume}{33}, \bibinfo{year}{2019},
  pp. \bibinfo{pages}{7208--7215}.
\bibitem[{Gururangan et~al.(2018)Gururangan, Swayamdipta, Levy, Schwartz,
  Bowman, and Smith}]{gururangan2018annotation}
\bibinfo{author}{S.~Gururangan}, \bibinfo{author}{S.~Swayamdipta},
  \bibinfo{author}{O.~Levy}, \bibinfo{author}{R.~Schwartz},
  \bibinfo{author}{S.~Bowman}, \bibinfo{author}{N.~A. Smith},
\newblock \bibinfo{title}{Annotation artifacts in natural language inference
  data},
\newblock in: \bibinfo{booktitle}{Proceedings of the 2018 Conference of the
  North American Chapter of the Association for Computational Linguistics:
  Human Language Technologies, Volume 2 (Short Papers)}, \bibinfo{year}{2018},
  pp. \bibinfo{pages}{107--112}.
\bibitem[{Poliak et~al.(2018)Poliak, Naradowsky, Haldar, Rudinger, and
  Van~Durme}]{poliak2018hypothesis}
\bibinfo{author}{A.~Poliak}, \bibinfo{author}{J.~Naradowsky},
  \bibinfo{author}{A.~Haldar}, \bibinfo{author}{R.~Rudinger},
  \bibinfo{author}{B.~Van~Durme},
\newblock \bibinfo{title}{Hypothesis only baselines in natural language
  inference},
\newblock in: \bibinfo{booktitle}{Proceedings of the Seventh Joint Conference
  on Lexical and Computational Semantics}, \bibinfo{year}{2018}, pp.
  \bibinfo{pages}{180--191}.
\bibitem[{Camburu et~al.(2018)Camburu, Rockt{\"a}schel, Lukasiewicz, and
  Blunsom}]{camburu2018snli}
\bibinfo{author}{O.-M. Camburu}, \bibinfo{author}{T.~Rockt{\"a}schel},
  \bibinfo{author}{T.~Lukasiewicz}, \bibinfo{author}{P.~Blunsom},
\newblock \bibinfo{title}{e-{SNLI}: natural language inference with natural
  language explanations},
\newblock in: \bibinfo{booktitle}{Advances in Neural Information Processing
  Systems}, \bibinfo{year}{2018}, pp. \bibinfo{pages}{9539--9549}.
\bibitem[{Thorne et~al.(2019)Thorne, Vlachos, Christodoulopoulos, and
  Mittal}]{thorne2019generating}
\bibinfo{author}{J.~Thorne}, \bibinfo{author}{A.~Vlachos},
  \bibinfo{author}{C.~Christodoulopoulos}, \bibinfo{author}{A.~Mittal},
\newblock \bibinfo{title}{Generating token-level explanations for natural
  language inference},
\newblock in: \bibinfo{booktitle}{Proceedings of the 2019 Conference of the
  North American Chapter of the Association for Computational Linguistics:
  Human Language Technologies, Volume 1 (Long and Short Papers)},
  \bibinfo{year}{2019}, pp. \bibinfo{pages}{963--969}.
\bibitem[{Silva et~al.(2019)Silva, Freitas, and Handschuh}]{silva2019exploring}
\bibinfo{author}{V.~S. Silva}, \bibinfo{author}{A.~Freitas},
  \bibinfo{author}{S.~Handschuh},
\newblock \bibinfo{title}{Exploring knowledge graphs in an interpretable
  composite approach for text entailment},
\newblock in: \bibinfo{booktitle}{{AAAI}}, \bibinfo{year}{2019}.
\bibitem[{Pawlik and Augsten(2016)}]{pawlik2016tree}
\bibinfo{author}{M.~Pawlik}, \bibinfo{author}{N.~Augsten},
\newblock \bibinfo{title}{Tree edit distance: Robust and memory-efficient},
\newblock \bibinfo{journal}{Information Systems} \bibinfo{volume}{56}
  (\bibinfo{year}{2016}) \bibinfo{pages}{157--173}.
\bibitem[{Zhang and Shasha(1989)}]{zhang1989simple}
\bibinfo{author}{K.~Zhang}, \bibinfo{author}{D.~Shasha},
\newblock \bibinfo{title}{Simple fast algorithms for the editing distance
  between trees and related problems},
\newblock \bibinfo{journal}{SIAM journal on computing} \bibinfo{volume}{18}
  (\bibinfo{year}{1989}) \bibinfo{pages}{1245--1262}.
\bibitem[{Chen and Manning(2014)}]{chen2014fast}
\bibinfo{author}{D.~Chen}, \bibinfo{author}{C.~Manning},
\newblock \bibinfo{title}{A fast and accurate dependency parser using neural
  networks},
\newblock in: \bibinfo{booktitle}{Proceedings of the 2014 conference on
  empirical methods in natural language processing (EMNLP)},
  \bibinfo{year}{2014}, pp. \bibinfo{pages}{740--750}.
\bibitem[{Clark et~al.(2008)Clark, Fellbaum, and Hobbs}]{clark2008using}
\bibinfo{author}{P.~Clark}, \bibinfo{author}{C.~Fellbaum},
  \bibinfo{author}{J.~Hobbs},
\newblock \bibinfo{title}{Using and extending {W}ord{N}et to support
  question-answering},
\newblock in: \bibinfo{booktitle}{Proceedings of the 4th Global WordNet
  Conference (GWC'08)}, \bibinfo{year}{2008}.
\bibitem[{Herrera et~al.(2006)Herrera, Penas, and Verdejo}]{herrera2006textual}
\bibinfo{author}{J.~Herrera}, \bibinfo{author}{A.~Penas},
  \bibinfo{author}{F.~Verdejo},
\newblock \bibinfo{title}{Textual entailment recognition based on dependency
  analysis and {W}ord{N}et},
\newblock in: \bibinfo{booktitle}{Machine Learning Challenges. Evaluating
  Predictive Uncertainty, Visual Object Classification, and Recognising Textual
  Entailment}, \bibinfo{publisher}{Springer}, \bibinfo{year}{2006}, pp.
  \bibinfo{pages}{231--239}.
\bibitem[{Fellbaum(1998)}]{fellbaum1998wordnet}
\bibinfo{author}{C.~Fellbaum}, \bibinfo{title}{Word{N}et},
  \bibinfo{publisher}{Wiley Online Library}, \bibinfo{year}{1998}.
\bibitem[{Silva et~al.(2016)Silva, Handschuh, and
  Freitas}]{silva2016categorization}
\bibinfo{author}{V.~S. Silva}, \bibinfo{author}{S.~Handschuh},
  \bibinfo{author}{A.~Freitas},
\newblock \bibinfo{title}{Categorization of semantic roles for dictionary
  definitions},
\newblock in: \bibinfo{booktitle}{Cognitive Aspects of the Lexicon (CogALex-V),
  Workshop at COLING 2016}, \bibinfo{year}{2016}, pp.
  \bibinfo{pages}{176--184}.
\bibitem[{M\`{a}rquez et~al.(2008)M\`{a}rquez, Carreras, Litkowski, and
  Stevenson}]{marquez2008semantic}
\bibinfo{author}{L.~M\`{a}rquez}, \bibinfo{author}{X.~Carreras},
  \bibinfo{author}{K.~C. Litkowski}, \bibinfo{author}{S.~Stevenson},
\newblock \bibinfo{title}{Semantic role labeling: an introduction to the
  special issue},
\newblock \bibinfo{journal}{Computational linguistics} \bibinfo{volume}{34}
  (\bibinfo{year}{2008}) \bibinfo{pages}{145--159}.
\bibitem[{Silva et~al.(2018)Silva, Freitas, and Handschuh}]{silva2018building}
\bibinfo{author}{V.~S. Silva}, \bibinfo{author}{A.~Freitas},
  \bibinfo{author}{S.~Handschuh},
\newblock \bibinfo{title}{Building a knowledge graph from natural language
  definitions for interpretable text entailment recognition},
\newblock in: \bibinfo{booktitle}{Proceedings of the Eleventh International
  Conference on Language Resources and Evaluation (LREC 2018)},
  \bibinfo{year}{2018}.
\bibitem[{Manning et~al.(2014)Manning, Surdeanu, Bauer, Finkel, Bethard, and
  McClosky}]{manning2014stanford}
\bibinfo{author}{C.~D. Manning}, \bibinfo{author}{M.~Surdeanu},
  \bibinfo{author}{J.~Bauer}, \bibinfo{author}{J.~R. Finkel},
  \bibinfo{author}{S.~Bethard}, \bibinfo{author}{D.~McClosky},
\newblock \bibinfo{title}{The {S}tanford {C}orenlp natural language processing
  toolkit},
\newblock in: \bibinfo{booktitle}{{ACL} ({S}ystem {D}emonstrations)},
  \bibinfo{year}{2014}, pp. \bibinfo{pages}{55--60}.
\bibitem[{Mesnil et~al.(2015)Mesnil, Dauphin, Yao, Bengio, Deng, Hakkani-Tur,
  He, Heck, Tur, Yu et~al.}]{mesnil2015using}
\bibinfo{author}{G.~Mesnil}, \bibinfo{author}{Y.~Dauphin},
  \bibinfo{author}{K.~Yao}, \bibinfo{author}{Y.~Bengio},
  \bibinfo{author}{L.~Deng}, \bibinfo{author}{D.~Hakkani-Tur},
  \bibinfo{author}{X.~He}, \bibinfo{author}{L.~Heck}, \bibinfo{author}{G.~Tur},
  \bibinfo{author}{D.~Yu}, et~al.,
\newblock \bibinfo{title}{Using recurrent neural networks for slot filling in
  spoken language understanding},
\newblock \bibinfo{journal}{IEEE/ACM Transactions on Audio, Speech and Language
  Processing (TASLP)} \bibinfo{volume}{23} (\bibinfo{year}{2015})
  \bibinfo{pages}{530--539}.
\bibitem[{Turney and Pantel(2010)}]{turney2010frequency}
\bibinfo{author}{P.~D. Turney}, \bibinfo{author}{P.~Pantel},
\newblock \bibinfo{title}{From frequency to meaning: Vector space models of
  semantics},
\newblock \bibinfo{journal}{Journal of artificial intelligence research}
  \bibinfo{volume}{37} (\bibinfo{year}{2010}) \bibinfo{pages}{141--188}.
\bibitem[{Marelli et~al.(2014)Marelli, Menini, Baroni, Bentivogli, Bernardi,
  and Zamparelli}]{marelli2014sick}
\bibinfo{author}{M.~Marelli}, \bibinfo{author}{S.~Menini},
  \bibinfo{author}{M.~Baroni}, \bibinfo{author}{L.~Bentivogli},
  \bibinfo{author}{R.~Bernardi}, \bibinfo{author}{R.~Zamparelli},
\newblock \bibinfo{title}{A {SICK} cure for the evaluation of compositional
  distributional semantic models},
\newblock in: \bibinfo{booktitle}{{LREC}}, \bibinfo{year}{2014}, pp.
  \bibinfo{pages}{216--223}.
\bibitem[{Freitas et~al.(2014)Freitas, da~Silva, Curry, and
  Buitelaar}]{freitas2014distributional}
\bibinfo{author}{A.~Freitas}, \bibinfo{author}{J.~C.~P. da~Silva},
  \bibinfo{author}{E.~Curry}, \bibinfo{author}{P.~Buitelaar},
\newblock \bibinfo{title}{A distributional semantics approach for selective
  reasoning on commonsense graph knowledge bases},
\newblock in: \bibinfo{booktitle}{International Conference on Applications of
  Natural Language to Data Bases/Information Systems},
  \bibinfo{organization}{Springer}, \bibinfo{year}{2014}, pp.
  \bibinfo{pages}{21--32}.
\bibitem[{Faralli and Navigli(2013)}]{faralli2013java}
\bibinfo{author}{S.~Faralli}, \bibinfo{author}{R.~Navigli},
\newblock \bibinfo{title}{A java framework for multilingual definition and
  hypernym extraction},
\newblock in: \bibinfo{booktitle}{Proceedings of the 51st Annual Meeting of the
  Association for Computational Linguistics: System Demonstrations},
  \bibinfo{year}{2013}, pp. \bibinfo{pages}{103--108}.
\bibitem[{Freitas et~al.(2012)Freitas, Curry, and
  O'Riain}]{freitas2012distributional}
\bibinfo{author}{A.~Freitas}, \bibinfo{author}{E.~Curry},
  \bibinfo{author}{S.~O'Riain},
\newblock \bibinfo{title}{A distributional approach for terminological semantic
  search on the linked data web},
\newblock in: \bibinfo{booktitle}{Proceedings of the 27th Annual {ACM}
  Symposium on Applied Computing}, \bibinfo{organization}{{ACM}},
  \bibinfo{year}{2012}, pp. \bibinfo{pages}{384--391}.
\bibitem[{Magnini et~al.(2014)Magnini, Zanoli, Dagan, Eichler, Neumann, Noh,
  Pado, Stern, and Levy}]{magnini2014excitement}
\bibinfo{author}{B.~Magnini}, \bibinfo{author}{R.~Zanoli},
  \bibinfo{author}{I.~Dagan}, \bibinfo{author}{K.~Eichler},
  \bibinfo{author}{G.~Neumann}, \bibinfo{author}{T.-G. Noh},
  \bibinfo{author}{S.~Pado}, \bibinfo{author}{A.~Stern},
  \bibinfo{author}{O.~Levy},
\newblock \bibinfo{title}{The excitement open platform for textual inferences},
\newblock in: \bibinfo{booktitle}{ACL (System Demonstrations)},
  \bibinfo{year}{2014}, pp. \bibinfo{pages}{43--48}.

\end{thebibliography}
	
\end{document}